\documentclass[11pt, a4paper, copyright, gr]{google}

\usepackage[authoryear, sort&compress, round]{natbib}
\usepackage{graphicx}
\usepackage{booktabs}
\usepackage{multirow}
\usepackage{amsmath}
\usepackage{caption} 
\usepackage{xcolor} 
\usepackage{textcomp} 
\usepackage{soul} 
\usepackage{mdframed}
\usepackage{needspace}
\usepackage{enumitem}
\usepackage{longtable}

\usepackage{float}
\usepackage{ltablex} 
\usepackage{array}
\usepackage{xurl}

\keepXColumns

\title{Earth AI: Unlocking Geospatial Insights with Foundation Models and Cross-Modal Reasoning}

\reportnumber{} 

\correspondingauthor{$\{$shekelto, sshetty$\}$@google.com}

\uselogo{} 

\author[*,1]{Aaron Bell}
\author[*,1]{Amit Aides}
\author[*,1]{Amr Helmy}
\author[*,1]{Arbaaz Muslim}
\author[*,1]{Aviad Barzilai}
\author[*,1]{Aviv Slobodkin}
\author[*,1]{Bolous Jaber}
\author[*,1]{David Schottlander}
\author[*,1]{George Leifman}
\author[*,1]{Joydeep Paul}
\author[*,1]{Mimi Sun}
\author[*,1]{Nadav Sherman}
\author[*,1]{Natalie Williams}
\author[*,1]{Per Bjornsson}
\author[*,1]{Roy Lee}
\author[*,1]{Ruth Alcantara}
\author[*,1]{Thomas Turnbull}
\author[*,1]{Tomer Shekel}
\author[*,6]{Vered Silverman}
\author[*,1]{Yotam Gigi}
\author[1]{Adam Boulanger}
\author[1]{Alex Ottenwess}
\author[2]{Ali Ahmadalipour}
\author[1]{Anna Carter}
\author[2]{Behzad Vahedi}
\author[3]{Charles Elliott}
\author[2]{David Andre}
\author[1]{Elad Aharoni}
\author[1]{Gia Jung}
\author[3]{Hassler Thurston}
\author[1]{Jacob Bien}
\author[1]{Jamie McPike}
\author[1]{Jessica Sapick}
\author[1]{Juliet Rothenberg}
\author[1]{Kartik Hegde}
\author[3]{Kel Markert}
\author[1]{Kim Philipp Jablonski}
\author[2]{Luc Houriez}
\author[1]{Monica Bharel}
\author[1]{Phing VanLee}
\author[1]{Reuven Sayag}
\author[1]{Sebastian Pilarski}
\author[5]{Shelley Cazares}
\author[1]{Shlomi Pasternak}
\author[3]{Siduo Jiang}
\author[1]{Thomas Colthurst}
\author[3]{Yang Chen}
\author[1]{Yehonathan Refael}
\author[1]{Yochai Blau}
\author[1]{Yuval Carny}
\author[4]{Yael Maguire}
\author[1]{Avinatan Hassidim}
\author[1]{James Manyika}
\author[$\dag$,1]{Tim Thelin}
\author[$\dag$,1]{Genady Beryozkin}
\author[$\dag$,1]{Gautam Prasad}
\author[$\dag$,1]{Luke Barrington}
\author[$\dag$,1]{Yossi Matias}
\author[$\dag$,1]{Niv Efron}
\author[$\dag$,1]{Shravya Shetty}

\affil[*]{Core contributor listed alphabetically}
\affil[$\dag$]{Co-last author}
\affil[1]{Google Research}
\affil[2]{Google X}
\affil[3]{Google Cloud}
\affil[4]{Google Geo}
\affil[5]{Google Public Sector}
\affil[6]{Work done at Google via Qualitest}

\begin{abstract}
Geospatial data offers immense potential for understanding our planet. However, the sheer volume and diversity of this data along with its varied resolutions, timescales, and sparsity pose significant challenges for thorough analysis and interpretation. This paper introduces Earth AI, a family of geospatial AI models and agentic reasoning that enables significant advances in our ability to unlock novel and profound insights into our planet. This approach is built upon foundation models across three key domains---Planet-scale Imagery, Population, and Environment---and an intelligent Gemini-powered reasoning engine. We present rigorous benchmarks showcasing the power and novel capabilities of our foundation models and validate that when used together, they provide complementary value for geospatial inference and their synergies unlock superior predictive capabilities. To handle complex, multi-step queries, we developed a Gemini-powered agent that jointly reasons over our multiple foundation models along with large geospatial data sources and tools. On a new benchmark of real-world crisis scenarios, our agent demonstrates the ability to deliver critical and timely insights, effectively bridging the gap between raw geospatial data and actionable understanding.
\end{abstract}

\begin{document}

\maketitle

\section{Introduction}
Understanding signals from our planet and reasoning about their effect on our livelihoods has inspired human curiosity and innovation for millennia, from the earliest origins of folklore as a guide to natural wisdom, to the first applications of computer science for weather forecasting \citep{lynch2008origins}.

Decades of siloed geospatial data from satellites \citep{donlon2012global, drusch2012sentinel}, sensors \citep{gorelick2017google, hersbach2020era5, munoz2021era5}, and demographic records \citep{DataCommons2025, DataCommons2025b} have posed a significant challenge to cross-domain analysis. To address this, the field of Geospatial AI (GeoAI) \citep{iyer2025harnessing} has evolved from specialized models to general-purpose Foundation Models for Earth Observation \citep{zhu2024foundationsearthclimatefoundation}. This shift, driven by large-scale datasets \citep{jakubik2023foundation, wang2023skyscript} and benchmarks \citep{mai2024opportunities}, has culminated in agentic systems where Large Language Models (LLMs) reason over geospatial data, tested by new frameworks \citep{kao2025towards} and reasoning benchmarks \citep{yerramilli2025geochain, dihan2025mapeval}.

\begin{figure}[H]
    \centering
    \includegraphics[width=0.9\textwidth]{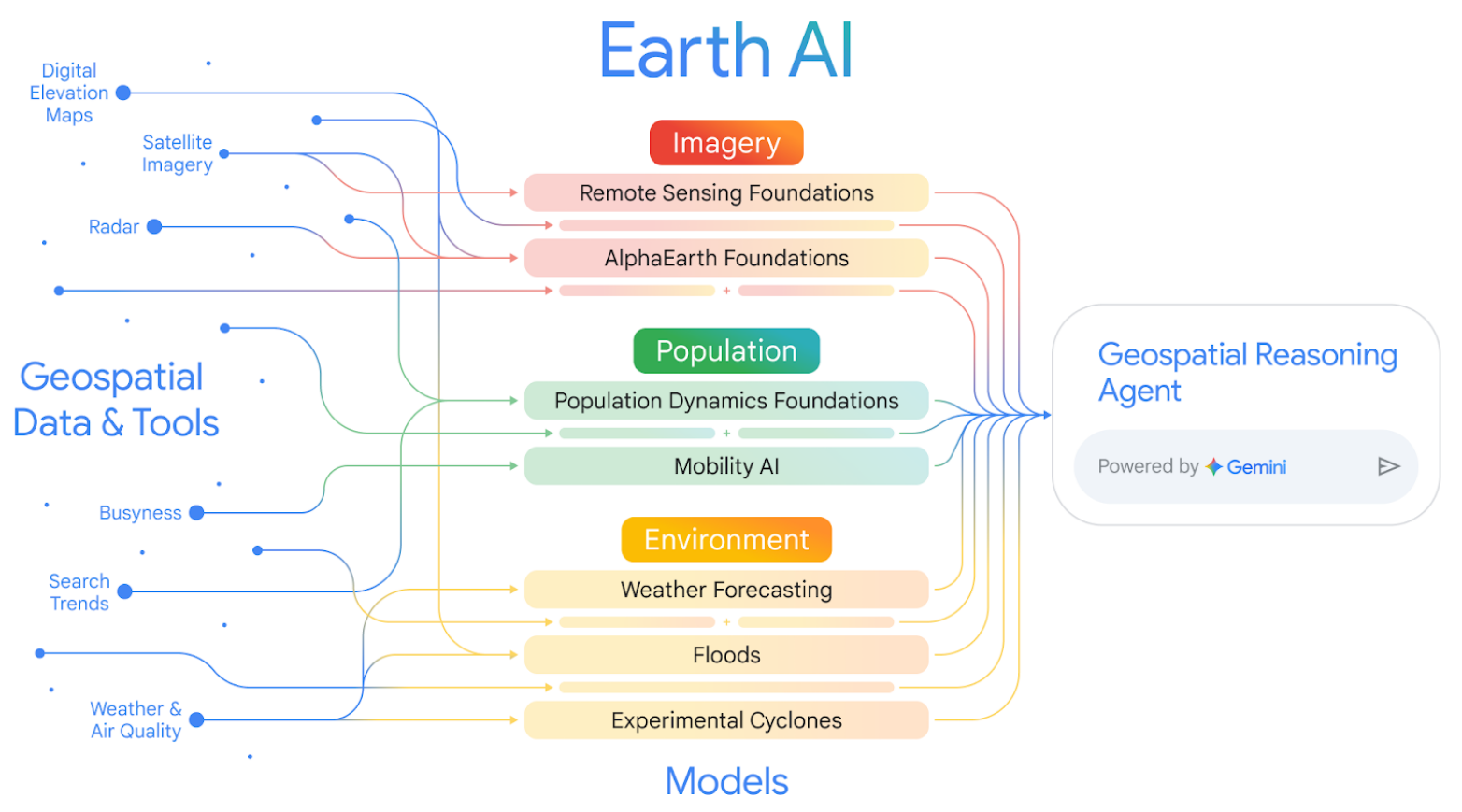} 
    \caption{EarthAI Overview: Earth AI builds on diverse geospatial data and tools across many sources and modalities (left) that are processed by specialized sub-agents and models spanning Imagery, Population, and Environment (middle). The Earth AI geospatial reasoning agent (right) ties these together to enable comprehensive geospatial analysis and insight generation.}
    \label{fig:overview}
\end{figure}

Our work builds on these foundations to propose a new paradigm for planetary analysis. We introduce ``Earth AI'', an interoperable family of geospatial AI models orchestrated by a customizable geospatial reasoning agent to create a holistic, multi-modal view of the Earth. Using Foundation Models (FMs) and LLM based reasoning, we build generalizable systems that surpass the limitations of single-purpose models and are capable of generating novel and actionable insights across a wide spectrum of planetary questions (see Figure~\ref{fig:overview} for an overview).

Our approach leverages three categories of Earth data: \textit{Imagery}, \textit{Population} and \textit{Environment}. For each category, we developed novel, foundation Earth AI Models that demonstrate state of the art performance in benchmark tasks.

\textit{Imagery} incorporates satellite, aerial and ground level imagery, sensor observations and related models \citep{gorelick2017google} of the planet, including mapping of urban and agricultural landscapes \citep{sirko2021continentalscalebuildingdetectionhigh, sirko2024highresolutionbuildingroaddetection, goroshin2023estimatingresidentialsolarpotential, dua2024agriculturallandscapeunderstandingcountryscale}, classifying global land cover \citep{brown2022dynamic}, discovering temporal urban patterns \citep{deng2025visualchroniclesusingmultimodal}, producing multi-source compact embeddings \citep{brown2025alphaearth} and the Remote Sensing Foundations vision backbones and multimodal models.

\textit{Population} encompasses observation and analysis of humans and their impact on the earth, including maps and data about the built environment \citep{weiss2020global}, simulation and optimization of mobility patterns and transportation systems (e.g., traffic, migration) \citep{aktay2020googlecovid19communitymobility, cook2025short, zhang2024traffic, choudhury2024towards, haddad2024quantitative}, demographic and socioeconomic patterns \citep{DataCommons2025} public health and vectors of disease transmission \citep{wellenius2021impacts} and our Population Dynamics Foundations integrating human behavior and location \citep{agarwal2024general}.

\textit{Environment} relates to spatiotemporal signals capturing dynamics of the Earth, including observations and models of weather, air quality, and climate \citep{agrawal2025operationaldeeplearningsatellitebased, lam2023learning, kochkov2024neural, price2025probabilistic, google2025weather, google2025airquality}, forecasts and tracking of natural disasters, such as cyclones \citep{alet2025skillful}, floods \citep{nearing2024global}, and wildfires \citep{matias2021realtime}, and maps of habitat loss and its underlying drivers \citep{sims2023global}.

To address real-world, multimodal challenges, we orchestrate these individual Earth AI Models with a Gemini-powered Geospatial Reasoning agent. By uniting natural language interactions and model connectivity across multiple domains that are usually analyzed separately, we expand the ability for non-expert users to analyze important questions without needing to download intermediate elements of answers and join or cross-reference results manually. This paradigm greatly expands the range of users to include anyone who can formulate a well-structured question and assess the resulting response. In this way, Earth AI allows for holistic, multi-faceted analysis and insight generation at a scale that was previously intractable. We demonstrate the power of this approach using a new benchmark of complex, real-world crisis response scenarios, showing how the synthesis of diverse data sources can unlock critical and time-sensitive insights.

The key findings from our evaluation of the Earth AI approach are summarized below:
\begin{itemize}
    \item \textbf{State-of-the-Art Earth AI Models:} We demonstrate that our Remote Sensing Foundations achieve state-of-the-art (SOTA) performance on tasks such as open-vocabulary object detection and zero-shot cross-modal retrieval. Concurrently, our Population Dynamics Foundations has been independently validated to improve real-world retail and public health applications, and has been extended to provide temporal embeddings at a monthly granularity.
    \item \textbf{Leading-edge Predictive Power Through Model Synergy:} We provide strong evidence that the integration of models from different modalities yields superior predictive capabilities. By combining signals from our Imagery, Population and Environment models and datasets, we achieve higher predictive accuracy on real-world classification and forecasting tasks versus a single modality analysis.
    \item \textbf{Complex Problem-Solving via Agentic Reasoning:} We show that the Gemini-powered reasoning agent can effectively deconstruct complex geospatial queries, select the appropriate models and tools in sequence, present transparent reasoning, and synthesize results into a coherent answer, demonstrating a robust capability to automate and scale complex analysis and insight generation across a number of domains, for example, identifying geographic regions with elevated short-term flood risk and high social vulnerability.
\end{itemize}

\section{Earth AI Capabilities}
We summarize the capabilities of Earth AI in this section. First, we introduce the core foundation Earth AI Models trained on specialized geospatial datasets. Next, we demonstrate how these models can be combined to create powerful predictive applications by leveraging their synergistic strengths. Finally, we describe an approach to orchestrating all of these components to solve complex, multi-step queries using agentic Geospatial Reasoning.

\subsection{Earth AI Models}
Our core Earth AI models are trained on specialized geospatial datasets across three categories of Earth data to analyze distinct aspects of our planet. First, we introduce our Imagery models, trained on remote sensing datasets. Next, we describe our Population model, which captures the dynamics of human behavior in relation to geography. Finally, we detail our suite of Environment models for weather, climate, and natural crisis applications.

\subsubsection{Imagery: Remote Sensing Foundations}
\begin{figure}[H]
    \centering
    \includegraphics[width=0.95\textwidth]{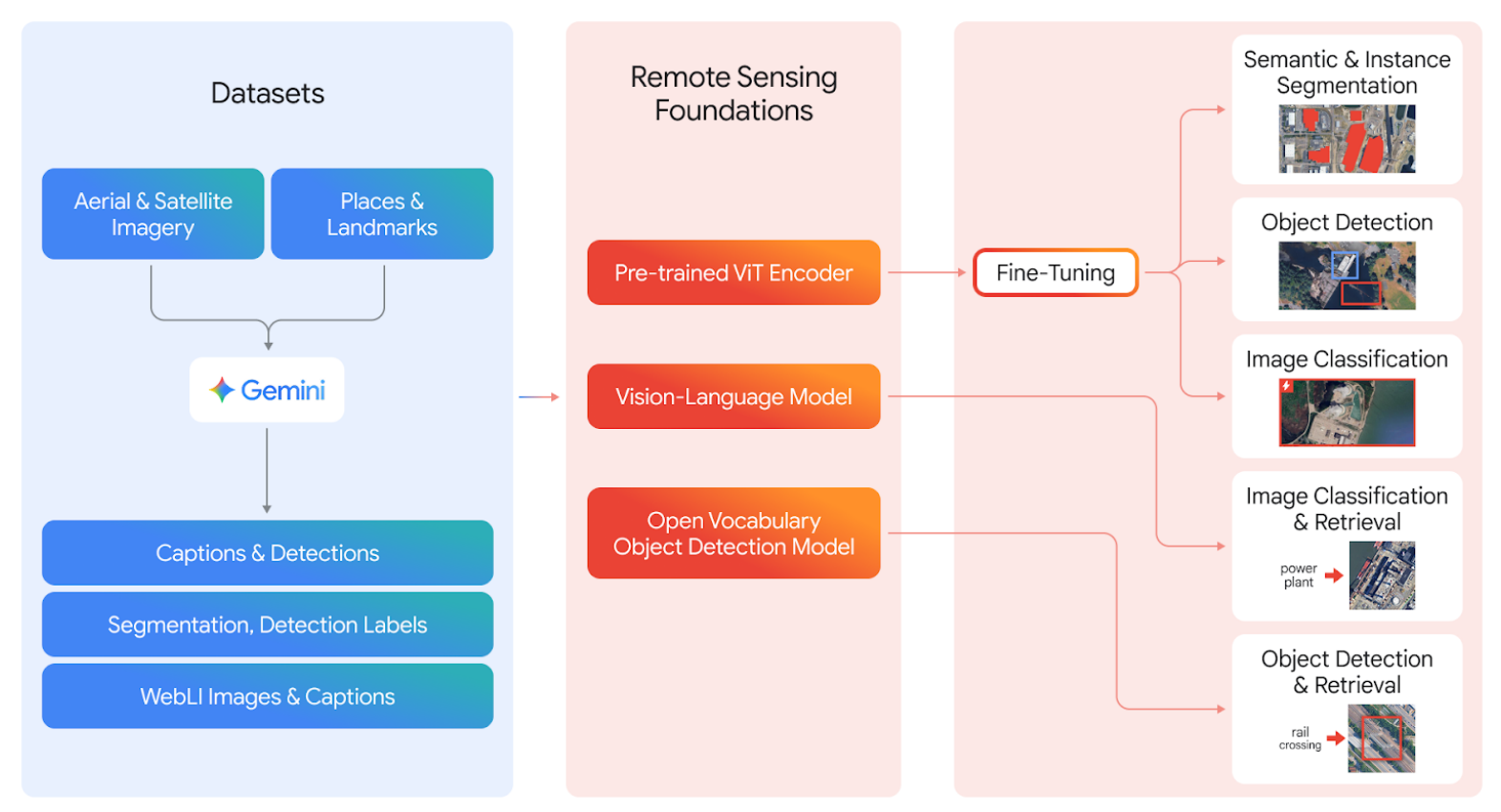} 
    \caption{Remote Sensing Foundation Models Overview: This diagram illustrates the training and use of RS Foundations models. Specialized aerial/satellite imagery and places data extracted from Google Maps are used to generate synthetic captions via Gemini. Together with text-annotated web imagery and annotated segmentation and detection datasets, they are used to train the Vision Transformer (ViT) Encoder, the Vision-Language Model, and the Open-Vocabulary Object Detection model. VLM and OVD models are then used directly on the respective classification, detection and retrieval tasks, and the ViT encoder is fine-tuned for optimal performance on downstream specific tasks.}
    \label{fig:rs_foundations}
\end{figure}

Existing generalist geospatial capabilities were built through large-scale pre-training on multispectral satellite imagery \citep{jakubik2023foundation} while specialized architectures like SatMAE utilize masked autoencoding for downstream tasks \citep{cong2022satmae}. Our Remote Sensing (RS) Foundation models (Figure~\ref{fig:rs_foundations}) address key challenges in Earth observation—such as limited labeled datasets and unique image distributions—to unlock new capabilities in visual understanding. These models provide a roadmap toward scalable, general-purpose RS analysis, bridging the gap between advancements in general computer vision and the specific demands of geospatial data.

This family of models features three core capabilities:
\begin{itemize}
    \item \textbf{Vision-Language Understanding:} We showcase vision-language models (VLMs) that connect remote sensing imagery with natural language. These models learn to map both visual and textual inputs into a joint embedding space, a process that enables the quantification of semantic similarity between an image and a corresponding text description.
    This core capability enables the model to perform dynamic, zero-shot image classification and retrieval using natural language prompts, effectively handling labels or descriptions not seen during its training.
    \item \textbf{Open-Vocabulary Object Detection:} We further present an open-vocabulary object detection (OVD) model that leverages VLM-derived embeddings. This allows the model to detect previously unseen object categories in a zero-shot setting, thus supporting detailed queries over satellite and aerial imagery. A Few-shot algorithm can further improve performance using just tens of annotated examples.
    \item \textbf{General-Purpose Vision-Transformer (ViT) Backbone:} We also feature a comprehensive method for pre-training a vision-transformer encoder on a combination of large-scale, unlabeled remote sensing imagery and smaller-scale labeled datasets. The resulting model was rigorously evaluated across scene classification, object detection, and semantic and instance segmentation tasks, demonstrating strong generalization capacity.
\end{itemize}

Our RS foundation models work in concert with \textit{AlphaEarth Foundations} \citep{brown2025alphaearth} to provide a multi-layered view of the planet. \textit{AlphaEarth Foundations} summarizes optical satellite images, radar, climate simulations and more, for macro-level analysis, reducing the need for large training datasets or directly handling satellite imagery. This is made available for analysis as a 10-meter resolution annual embedding. The Remote Sensing Foundations complement the offering by providing direct access to models that operate on RGB imagery from diverse sources at fine-grained resolutions (0.1m-10m), and can be fine-tuned for specific tasks. Additionally Remote Sensing Foundations models feature native support for natural language queries, enabling non-experts to conduct rapid analysis of imagery that captures specific objects or events.

\subsubsection{Population: Population Dynamics Foundations}
\begin{figure}[H]
    \centering
    \includegraphics[width=0.95\textwidth]{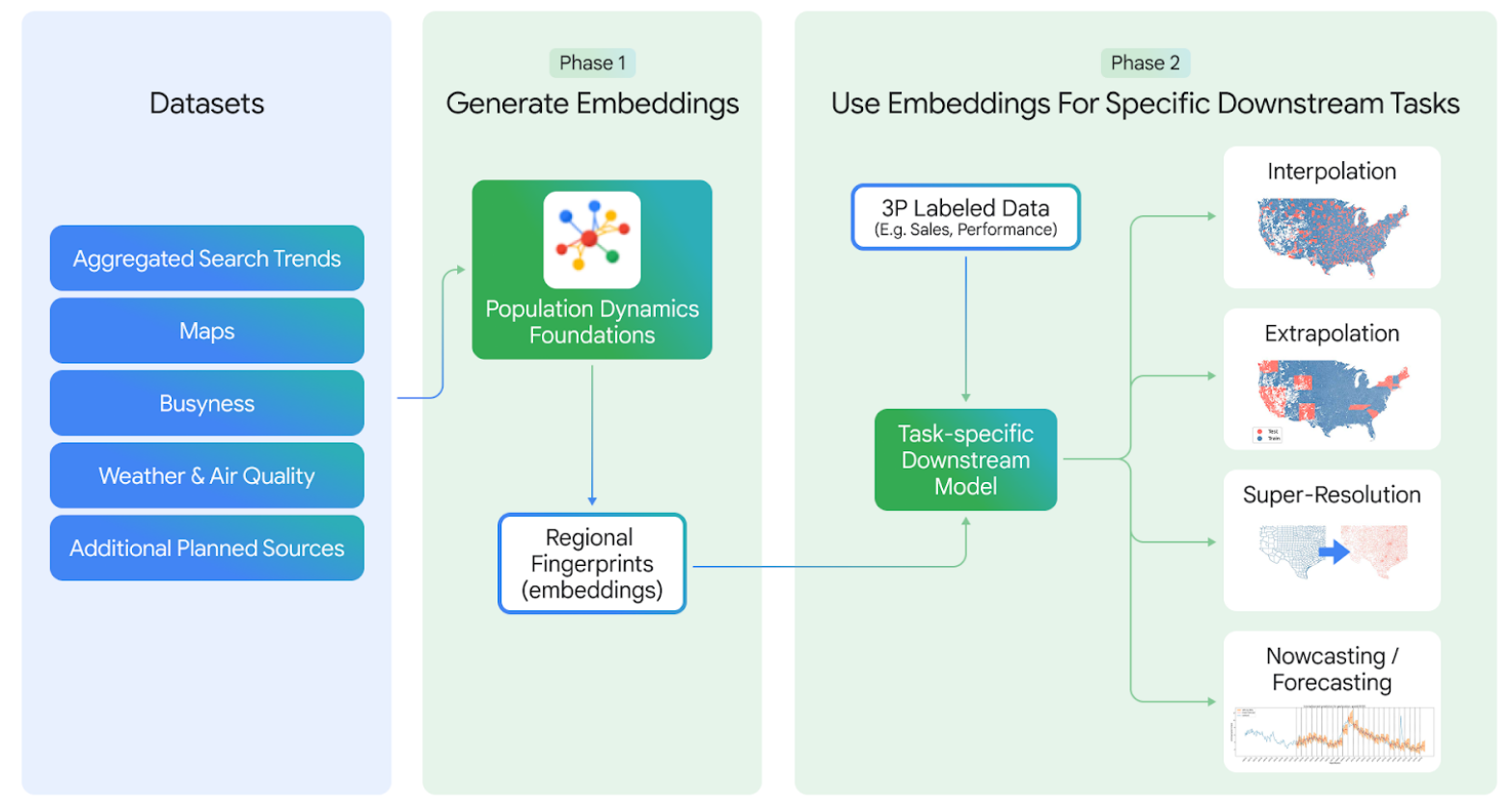} 
    \caption{Population Dynamics Foundations Two-Phase Framework: Phase 1 involves offline training using large diverse geospatial data (Maps, Search Trends, Busyness and environmental conditions) to encode compact Regional embeddings. Phase 2 uses these pre-trained embeddings to enable dynamic fine-tuning for granular downstream tasks, including interpolation, extrapolation, super-resolution, and forecasting of local statistics.}
    \label{fig:pdf_framework}
\end{figure}

Our \textit{Population Dynamics Foundations} model fuses diverse datasets to represent the dynamics of human behavior in a geographic context \citep{agarwal2024general}, as presented in Figure~\ref{fig:pdf_framework}. It captures the built environment through maps data, the natural environment through weather and air quality, and human behavior through Search Trends and anonymized busyness data. The model relates and encodes these datasets in a graph neural network to produce a unified digital embedding for each region (e.g., administrative region, postal code) while preserving privacy.

Our original \textit{Population Dynamics Foundations} focused on US data over a single year \citep{agarwal2024general}. This work builds on that with two key expansions for enhanced analytical capabilities:
\begin{itemize}
    \item \textbf{Global Spatial Coverage:} We have expanded embeddings to 17 countries. The resulting Global Population Dynamics Foundations embeddings are comparable across countries, meaning a downstream model trained on US data can be applied in the United Kingdom or Brazil.
    \item \textbf{Dynamic Temporal Analysis:} We have created embeddings that evolve over time, represented as monthly embeddings over the last two years. This extension directly addresses the challenge of unpredictable human behavior by integrating temporal behavioral shifts into the embeddings, enhancing nowcasting and forecasting applications.
\end{itemize}

\subsubsection{Environment: Weather \& Climate Models}
Our Environment models and APIs provide state-of-the-art insights into weather, climate, air quality and natural crises, making complex geospatial information widely available. In this evaluation, we have integrated three distinct, representative environmental signals:
\begin{itemize}
    \item \textbf{Weather Forecasting:} Google Maps Platform Weather API, incorporates machine learning models such as MetNet \citep{agrawal2025operationaldeeplearningsatellitebased}, and provides hourly forecasts up to 240 hours and daily forecasts up to 10 days, covering conditions such as temperature, precipitation, wind, and UV index.
    \item \textbf{Flood Forecasting:} The Google Flood Forecasting API delivers real-time riverine flood predictions using data from measurement gauges. Forecasts detail the anticipated area of inundation, severity level, and probability. Users can retrieve current predictions or access historical forecast data back to August 1st, 2025.
    \item \textbf{Experimental Cyclone Forecasting:} Google's experimental AI-based cyclone model, based on stochastic neural networks \citep{alet2025skillful}, predicts a cyclone's formation, track, intensity, size and shape by generating 50 possible scenarios up to 15 days in advance. Historical data is available back to January 1st, 2022.
\end{itemize}

\subsection{Combining Earth AI Models: Predictive Applications}
While each Earth AI model offers a powerful lens into a specific domain, a holistic understanding of our planet requires us to leverage multiple domains simultaneously. Any single model, whether focused on imagery, human behavior, or climate, is inherently limited by its perspective. Earth AI is designed to overcome this limitation by allowing for the integration of these diverse viewpoints, enabling a more comprehensive analysis than would otherwise be possible. For instance, embeddings like \textit{AlphaEarth Foundations} and \textit{Population Dynamics Foundations} provide complementary, location-specific representations: \textit{Population Dynamics} indexes human-centric signals such as Search Trends, mobility and maps while \textit{AlphaEarth Foundations} encodes imagery, topography and climate information to give structural and environmental context.

We describe a number of methods for combining Earth Models by first mapping them to the same administrative regions and then integrating representations into geospatial modeling tasks such as extrapolation and forecasting.

\subsection{Orchestration: Solving Complex Queries with Geospatial Reasoning Agents}
The ultimate goal of Earth AI is to help users answer complex, real-world questions that require multifaceted reasoning across diverse models and data sources. Such queries can be categorized into a hierarchy of increasing complexity:
\begin{enumerate}
    \item Descriptive and retrieval queries involving fact-finding (e.g., \textit{``What was the highest recorded temperature in New York in August 2020?''}).
    \item Analytical and relational queries seeking to uncover patterns between different data sources (e.g., \textit{``How many hospitals were located in areas experiencing severe storm conditions in the state of Louisiana when Hurricane Katrina came ashore?''}).
    \item Predictive or inferential queries involving forecasting new information (e.g., \textit{``Which Indian cities have the most vulnerable populations at high risk of being impacted by flooding by November 25, 2027?''}).
\end{enumerate}

\begin{figure}[H]
    \centering
    \includegraphics[width=\textwidth]{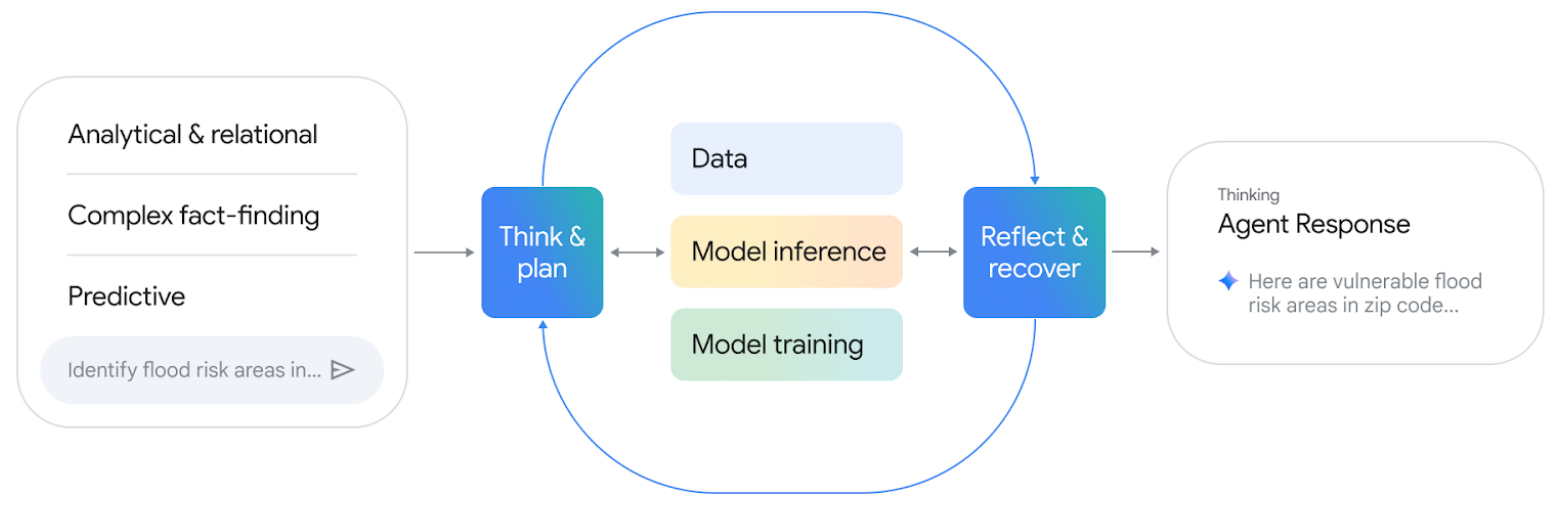} 
    \caption{Geospatial Reasoning Agent Operational Framework: This closed-loop process is designed to solve queries across three complexity categories (Analytical, Fact-finding, Predictive). The agent iteratively refines its response through the cycle of Think \& Plan, Data Operations/Model Inference/Training, and Reflect \& Recover until a final, grounded response is produced.}
    \label{fig:agent_framework}
\end{figure}

While direct data retrieval is sufficient for simple queries, addressing analytical and predictive questions demands a more sophisticated methodology. To meet this challenge, we developed the Geospatial Reasoning Agent (Figure~\ref{fig:agent_framework}). Its main role is to serve as an intelligent intermediary, both planning tasks and connecting to a nuanced understanding of the world derived from our core Earth Models and datasets. 
The agent is designed to decompose complex problems, leveraging its specialized capabilities to interpret Earth observation data with relevant Imagery models, infer socio-demographic insights using \textit{Population Dynamics Foundations}, and generate forecasts with our \textit{Environment models}.

The Geospatial Reasoning Agent, developed with Google's Agent Development Kit (ADK) and powered by Gemini, integrates general-purpose capabilities like orchestration, planning, and recovery with specialized geospatial functionalities.
These specialized capabilities are domain-specific, implemented as either simple tools or complex `expert' sub-agents, based on task complexity.
This modularity is aimed to facilitate future extensibility and customization. 
We categorize these capabilities into four primary domains, alongside more general functions:
\begin{description}
    \item[Imagery:] Utilizes the \textit{RemoteSensing Foundations} family of models to perform on-demand analysis of satellite imagery, including tasks like classification, object detection and retrieval.
    \item[Population:] Leverages our \textit{Population Dynamics Foundations} models, alongside Google's Places API and Data Commons, to resolve geographic boundaries and provide dynamic demographic statistics.
    \item[Environment:] Accesses and reasons over dynamic Earth processes, using historical atmospheric data and our predictive models for cyclones, floods, and other phenomena.
    \item[Spatiotemporal Model Training:] Provides a natural language interface for on-the-fly model training, using pre-trained embeddings from our core models to perform predictive tasks for user-specified variables.
    \item[General Capabilities:] The system is augmented with essential tools for geospatial data analysis, code-generation for custom analyses, Google Earth Engine for access to public and private geospatial datasets, Google Search as an additional knowledge source, and key Google Cloud services.
\end{description}

Accessible through a natural language, map-based user interface, the agent interprets a user's query, breaks it into manageable sub-tasks, delegates each to the appropriate expert agent or tool, and synthesizes the final response. This synergetic relationship—where capable models distill reality and a powerful agent reasons over that distillation—enables both retrospective investigation and proactive planning for complex scenarios.

\section{Evaluation and Results}
In this section, we present a comprehensive evaluation of the Earth AI capabilities. We first validate the performance of our new foundation models for Imagery and Population individually, demonstrating state-of-the-art results on established benchmarks. We then show the synergistic power of combining these models for complex predictive tasks. Finally, we evaluate the capabilities of our Geospatial Reasoning Agent on a new benchmark of real-world analytical and crisis-response queries.

\subsection{Earth AI Models}
\subsubsection{Imagery: Remote Sensing Foundations}
The Remote Sensing Foundations are composed of three core capabilities: a contrastively-trained VLM for tasks like zero-shot classification and cross-modal retrieval, an open-vocabulary object detection model for identifying objects without predefined labels, and a versatile pre-trained vision backbone model that can be fine-tuned for various downstream tasks. In this section, we present the evaluation results for each of these components.

\paragraph{Remote Sensing Vision Language Models}
We evaluated our VLMs on two key tasks: zero-shot classification and text-based retrieval, using several public remote sensing benchmarks. Evaluations (Table~\ref{tab:zsc_results} and Table~\ref{tab:retrieval_results}) show that our training datasets applied to the SigLIP2 and MaMMUT architectures result in state-of-the-art model performance on the vast majority of benchmarks and are comparable with much larger and more expensive chat-based models like GeoChat (7B) \citep{kuckreja2024geochat} and LHRS-Bot (7B) \citep{muhtar2024lhrs}. Additional details can be found in \citep{barzilai2025recipe}.

\begin{table}[H]
\centering
\caption{Remote Sensing zero-shot classification: Top-1 accuracy for zero-shot classification performance. Our results (RS MaMMUT and RS SigLIP2) are compared to SOTA VLM models and to the results achieved using baseline SigLIP2 and MaMMUT models. Bold indicates the highest value per benchmark (see footnote for SkyScript).}
\label{tab:zsc_results}
\resizebox{\textwidth}{!}{%
\begin{tabular}{lccccc}
\toprule 
& \begin{tabular}[c]{@{}c@{}}FMOW \\ \citep{christie2018fmow}\end{tabular} & \begin{tabular}[c]{@{}c@{}}SkyScript \\ \citep{wang2023skyscript}\end{tabular} & \begin{tabular}[c]{@{}c@{}}RESISC45 \\ \citep{Cheng2017RESISC45}\end{tabular} & \begin{tabular}[c]{@{}c@{}}UCM \\ \citep{qu2016deep}\end{tabular} & \begin{tabular}[c]{@{}c@{}}AID \\ \citep{xia2017aid}\end{tabular} \\ \midrule
SkyScript \citep{wang2023skyscript} & 28.04 & 70.89* & 70.94 & -- & -- \\
RS-CLIP \citep{li2023rs} & -- & 68.84 & 71.35 & 74.28 & 70.51 \\
GeoRSCLIP (VitL) \citep{zhang2024rs5m} & -- & -- & 71.89 & -- & 76.33 \\
RemoteCLIP \citep{liu2024remoteclip} & -- & -- & 79.84 & -- & \textbf{91.30} \\
MaMMUT 400M \citep{kuo2023mammut} & 37.58 & 58.66 & 66.93 & 76.52 & 71.46 \\
SigLIP2 400M \citep{tschannen2025siglip} & 41.25 & 65.94 & 72.40 & 81.43 & 75.56 \\
\midrule
RS-MaMMUT 400M (ours) & 47.24 & \textbf{69.46} & 72.31 & 80.29 & 71.96 \\
RS-SigLIP2 400M (ours) & \textbf{48.13} & 68.13 & \textbf{80.13} & \textbf{84.86} & 78.26 \\ 
\bottomrule
\end{tabular}%
}
\captionsetup{labelsep=none}
\caption*{{\tiny * SkyScript model evaluation was on an ``in-domain'' test set, thus an easier task compared to models and not directly comparable to the other models.}}
\end{table}

\begin{table}[H]
\centering
\caption{Remote Sensing zero-shot retrieval: Average of top-1/5/10 of zero-shot retrieval results for image to text (I2T) and text to image(T2I). Our results (RS MaMMut and RS SigLIP2) are compared to three SOTA VLM models and to the results achieved using baseline SigLIP2 and MaMMUT models. Bold indicates the highest value per benchmark.}
\label{tab:retrieval_results}
\resizebox{\textwidth}{!}{%
\begin{tabular}{l c@{\hskip 30pt}c c@{\hskip 30pt}c c@{\hskip 35pt}c c@{\hskip 40pt}c}
\toprule
& \multicolumn{2}{c}{RSICD} & \multicolumn{2}{c}{UCM-Captions} & \multicolumn{2}{c}{RSITMD} & \multicolumn{2}{c}{NWPU} \\
& \multicolumn{2}{c}{\citep{lu2017exploring}} & \multicolumn{2}{c}{\citep{qu2016deep}} & \multicolumn{2}{c}{\citep{yuan2022exploring}} & \multicolumn{2}{c}{\citep{Cheng2022NWPU}} \\
\cmidrule(lr){2-3} \cmidrule(lr){4-5} \cmidrule(lr){6-7} \cmidrule(lr){8-9}
& I2T & T2I & I2T & T2I & I2T & T2I & I2T & T2I \\ \midrule
PIR-ITR \citep{pan2024pir} & 24.43 & 25.77 & -- & -- & 38.64 & 39.85 & -- & -- \\
SkyScript SkyCLIP-30 \citep{wang2023skyscript} & 23.70 & 19.97 & 72.22 & 59.33 & 30.75 & 30.58 & -- & -- \\
Geo-RSClip+RS5M \citep{zhang2024rs5m} & 26.41 & 25.96 & -- & -- & 33.33 & 38.02 & -- & -- \\
MaMMUT 400M \citep{kuo2023mammut} & 23.88 & 24.17 & 69.21 & 66.50 & 28.83 & 32.70 & 20.33 & 23.18 \\
SigLIP2 400M \citep{tschannen2025siglip} & 27.36 & 27.93 & 70.48 & 68.52 & 24.62 & 25.14 & 31.86 & 36.70 \\
\midrule
RS-MaMMut 400M (ours) & 33.33 & 33.59 & 74.76 & 71.79 & 42.63 & 42.58 & 41.44 & 32.28 \\
RS-SigLIP2 400M (ours) & \textbf{38.37} & \textbf{37.64} & \textbf{76.67} & \textbf{75.33} & \textbf{43.14} & \textbf{47.26} & \textbf{45.12} & \textbf{37.74} \\ \bottomrule
\end{tabular}%
}
\end{table}

\paragraph{Remote Sensing Open Vocabulary Detection}
We assessed our Remote Sensing (RS) Open-Vocabulary Detection (OVD) model's zero-shot performance on two commonly-used remote sensing object detection datasets: DOTA \citep{xia2018dota} and DIOR \citep{dior2023rs}. As shown in Table~\ref{tab:ovd_results} (top), our RS-OWL-ViT-v2 OVD model outperforms the baseline OWL-ViT-v2 on both benchmarks, achieving mean Average Precision (mAP) scores of 31.83\% on DOTA and 29.39\% on DIOR.

We also evaluated our OVD model when augmented with a few-shot learning technique, following the FLAME approach \citep{refael2025ontheflyovdadaptationflame} (Table~\ref{tab:ovd_results}, bottom). Using just 30 labeled classification examples per category, this approach improved the mAP to 53.96\% on DOTA and 53.21\% on DIOR, thereby demonstrating its effectiveness in adapting the model with minimal additional data. 
Notably, this variant also outperformed the recently-proposed Scale-adaptive Intersection over Union (SIoU) \citep{jeune2023SIoU} approach for few-shot object detection, further confirming its utility.

\begin{table}[H]
\centering
\caption{Remote Sensing Open Vocabulary Detection for Zero-shot and Few-shot: Comparing the mAP performance of the baseline OWL-ViTv2 open vocabulary detection model and our remote sensing variant on public object detection datasets for the Zero-shot and Few-shot regimes. Bold indicates the highest value per benchmark and setting (zero or few shot).}
\label{tab:ovd_results}
\begin{tabular}{lcc}
\toprule
Model & DOTA & DIOR \\ \midrule
\multicolumn{3}{l}{\textit{Zero-Shot}} \\
\hspace{3mm} {\small OWL-ViT-v2 \citep{minderer2023scaling}} & 13.77\% & 14.98\% \\
\hspace{3mm} {\small RS-OWL-ViT-v2 (ours)} & \textbf{31.83\%} & \textbf{29.39\%} \\ \midrule
\multicolumn{3}{l}{\textit{Few-Shot}} \\
\hspace{3mm} {\small SIoU \citep{jeune2023SIoU}} & 45.88\% & 52.85\% \\
\hspace{3mm} {\small FLAME with RS-OWL-ViT-v2 (ours) } & \textbf{53.96\%} & \textbf{53.21\%} \\ \bottomrule
\end{tabular}
\end{table}

\paragraph{Remote Sensing Pre-trained Backbone Foundation Model}
We evaluated the performance of our remote sensing global multi-task pretraining (RS-Global MTP) backbone model on 13 different downstream fine-tuning benchmarks from four categories:
\begin{itemize}
    \item \textbf{Image Classification:} FMoW \citep{christie2018fmow}, Resisc45 \citep{Cheng2017RESISC45}, UCM, AID \citep{xia2017aid}, SKAI \citep{lee2020assessing}
    \item \textbf{Object Detection:} DIOR \citep{dior2023rs}
    \item \textbf{Semantic Segmentation:} FloodNet \citep{rahnemoonfar2021floodnet}, FLAIR \citep{garioud2023flair}, DLRSD \citep{shao2018performance}, WHDLD \citep{shao2020multilabel}, Farmed Landscape \citep{conserva2025mapping}, SpaceNet7 \citep{van2021multi}
    \item \textbf{Instance Segmentation} Open Buildings \citep{sirko2021continentalscalebuildingdetectionhigh} and Agricultural Landscape Understanding \citep{dua2024agriculturallandscapeunderstandingcountryscale}
\end{itemize}

\textbf{Comparison to Baselines:} As detailed in Table~\ref{tab:backbone_comparison}, our model demonstrates superior overall performance compared to other leading backbone models like Dino V2 and SigLIP2. Our RS-Global MTP model achieved the highest average improvement of 14.93\% over the standard ImageNet pre-trained baseline across all task categories. All models were evaluated with tuneable backbone weights, the same batch size and for the same number of epochs, while choosing the optimal learning rate for each model.

\textbf{State-of-the-Art Performance:} We also compared our model against the current remote sensing state-of-the-art (SOTA) on public benchmarks with established data splits. As seen in Table~\ref{tab:sota_comparison}, our RS-Global MTP model surpasses previous SOTA results in classification on FMoW (81.70\%), segmentation on FLAIR (65.72\%), and object detection on DIOR (85.50\%).

\begin{table}[H]
\centering
\caption{Remote Sensing Pre-trained Foundation Mode: Results from fine-tuning our Remote Sensing pre-trained backbone for semantic segmentation, object detection, classification, Instance Segmentation tasks, compared to commonly used backbones. Comparison was performed using identical training configurations, except learning rate which was tuned for each model. All results are relative to the ViTL/16 pretrained on ImageNet. Bold indicates the highest value per benchmark.}
\label{tab:backbone_comparison}
\resizebox{\textwidth}{!}{%
\begin{tabular}{lcccc}
\toprule
Backbone model & \begin{tabular}[c]{@{}c@{}}Semantic segmentation \\ (mIoU)\end{tabular} & \begin{tabular}[c]{@{}c@{}}Object Detection \\ (AP50)\end{tabular} & \begin{tabular}[c]{@{}c@{}}Classification \\ (accuracy)\end{tabular} & \begin{tabular}[c]{@{}c@{}}Instance Seg \\ (mAP)\end{tabular} \\ \midrule
Random ViTL/16 & -16.23\% & -17.01\% & -24.14\% & -40.87\% \\
ViTL/16 pretrained on ImageNet (baseline) & 0.00\% & 0.00\% & 0.00\% & 0.00\% \\
Dino V2 ViTL/14 \citep{oquab2023dinov2} & 4.44\% & 4.69\% & \textbf{4.89\%} & 75.94\% \\
SigLIP2 So400m14 \citep{tschannen2025siglip} & 4.50\% & 2.47\% & 4.74\% & 63.75\% \\
RS SigLIP2 So400m14 \citep{barzilai2025recipe} & 4.15\% & 2.52\% & 4.79\% & 66.74\% \\
RS-Global MTP (ours) & \textbf{6.09\%} & \textbf{5.91\%} & 4.58\% & \textbf{82.92\%} \\ \bottomrule
\end{tabular}%
}
\end{table}

\begin{table}[H]
\centering
\caption{State-of-the-art comparison of evaluated foundation models using public datasets (classification, segmentation and object detection): For a fair comparison we compare only the downstream tasks for which performance was reported on publicly available train/test data splits. Bold indicates the highest value per benchmark.}
\label{tab:sota_comparison}
\resizebox{\textwidth}{!}{%
\begin{tabular}{lccc}
\toprule
& \textbf{Classification (Accuracy)} & \textbf{Segmentation (mIoU)} & \textbf{Object Detection (AP@50)} \\
& FMoW \citep{christie2018fmow} & FLAIR \citep{garioud2023flair} & DIOR \citep{dior2023rs} \\ \midrule
SOTA RS foundation model & 79.30\% \citep{khanna2024explora_fmow} & 63.10\% \citep{dimitrovski2025deep_flair} & 81.10\% \citep{wang2024mtp_dior} \\
RS-Global MTP (ours) & \textbf{81.70\%} & \textbf{65.72\%} & \textbf{85.50\%} \\ \bottomrule
\end{tabular}%
}
\end{table}

\subsubsection{Population: Population Dynamics Foundations}
Having established the state-of-the-art performance of our models for interpreting the physical and built environment using Remote Sensing Foundation, we now turn to the human dimension. We’ve previously shown that the embeddings derived from \textit{Population Dynamics Foundations} \citep{agarwal2024general} can power a range of downstream geospatial tasks including interpolation, extrapolation, super-resolution, and when combined with the \textit{TimesFM} model \citep{das2024decoderonly} also forecasting of socio-demographic data. In that work, we compare \textit{Population Dynamics Foundations} to existing foundation models using 29 target variables covering health, socioeconomic and environmental categories from Google Data Commons and Google Earth Engine (see the full list in Appendix~\ref{subsubsec:appendix_population_dynamics_foundations}). We found that when we compared with existing foundation models like remote sensing based \textit{SatCLIP} \citep{klemmer2025satclip} or personal photography based \textit{GeoCLIP} \citep{cepeda2023geoclip}, \textit{Population Dynamics Foundations} was able to achieve state of the art performance on most target variables and tasks.

In this section, we provide performance metrics on the newly developed global embeddings, temporal embeddings and results from independent external validation.

\paragraph{Global Population Dynamics Embeddings}
Table~\ref{tab:global_pdf_results} demonstrates the performance of the Global Population Dynamics Foundations on spatial interpolation tasks where the modeling problem is to predict missing target variables across 20\% of the administrative regions in each country. We used $R^2$ as our metric to assess performance (higher is better). This analysis focuses on four target globally available variables – elevation, night-time lights, population density, and tree cover – retrieved from Google Earth Engine.

\begin{table}[H]
\centering
\caption{Global Population Dynamics Foundations embeddings for 17 countries: $R^2$ score by country for using Global Population Dynamics Foundations embeddings to predict 4 target variables across 17 countries. For each target variable and country, a model is trained on labels from 80\% of the regions in the country to predict for the remaining 20\% of regions. Bold indicates the mean value across benchmarks.}
\label{tab:global_pdf_results}
\resizebox{0.7\textwidth}{!}{%
\begin{tabular}{c c c c c c}
\toprule
Country & Elevation & Night-time & Population & Tree Cover & Mean \\
Code & & Lights & Density & (\%) & \\
\midrule
AU & 0.89 & 0.93 & 0.92 & 0.79 & 0.88 \\
BE & 0.93 & 0.91 & 0.90 & 0.64 & 0.89 \\
BR & 0.96 & 0.90 & 0.87 & 0.85 & 0.89 \\
CA & 0.93 & 0.87 & 0.85 & 0.79 & 0.86 \\
CH & 0.89 & 0.82 & 0.85 & 0.50 & 0.77 \\
DE & 0.95 & 0.92 & 0.94 & 0.52 & 0.83 \\
ES & 0.95 & 0.91 & 0.93 & 0.82 & 0.90 \\
FR & 0.96 & 0.91 & 0.92 & 0.67 & 0.87 \\
GB & 0.81 & 0.92 & 0.89 & 0.65 & 0.82 \\
IN & 0.97 & 0.90 & 0.87 & 0.85 & 0.90 \\
IT & 0.93 & 0.90 & 0.89 & 0.82 & 0.89 \\
JP & 0.91 & 0.94 & 0.94 & 0.81 & 0.90 \\
MX & 0.98 & 0.91 & 0.89 & 0.87 & 0.91 \\
NG & 0.79 & 0.52 & 0.52 & 0.71 & 0.63 \\
NL & 0.87 & 0.88 & 0.89 & 0.42 & 0.77 \\
PT & 0.83 & 0.92 & 0.94 & 0.82 & 0.88 \\
US & 0.98 & 0.93 & 0.88 & 0.87 & 0.92 \\
\midrule
All Countries Average & 0.91 & 0.88 & 0.88 & 0.73 & 0.85 \\
\bottomrule
\end{tabular}%
}
\end{table}

We also used a set of variables from Eurostat at the NUTS 3 region level to evaluate the embedding's ability to generalize across countries. Data from Belgium, Switzerland, Germany, Spain, United Kingdom, Italy and Netherlands are used to train a model to make predictions for France. The results are shown in Table~\ref{tab:cross_country_extrapolation}.

\begin{table}[H]
\centering
\caption{Cross-country extrapolation with Global Population Dynamics Foundations embeddings: This table uses the NUTS3 region-level data in Belgium, Switzerland, Germany, Spain, United Kingdom, Italy, and Netherlands to make predictions for a new country - France. Two commonly used spatial interpolation methods are used as baselines for comparison: Inverse Distance Weighting (IDW) and Radial Basis Functions (RBF). Bold indicates the best value per benchmark (highest for $R^2$, lowest for MAPE).}
\label{tab:cross_country_extrapolation}
\resizebox{\textwidth}{!}{%
\begin{tabular}{l cc cc cc}
\toprule
& \multicolumn{2}{c}{GDP per capita} & \multicolumn{2}{c}{Death rate} & \multicolumn{2}{c}{Fertility rate} \\
\cmidrule(lr){2-3} \cmidrule(lr){4-5} \cmidrule(lr){6-7}
Model & $R^2 \uparrow$ & MAPE (\%) $\downarrow$ & $R^2 \uparrow$ & MAPE (\%) $\downarrow$ & $R^2 \uparrow$ & MAPE (\%) $\downarrow$ \\ \midrule
Inverse Distance Weighting & 0.00 & 28.03 & -0.07 & 19.29 & 0.51 & 8.78 \\
Radial Basis Functions & 0.13 & 25.71 & 0.52 & 12.91 & 0.50 & 9.35 \\
PopulationDynamics Foundations & \textbf{0.52} & \textbf{19.03} & \textbf{0.78} & \textbf{7.99} & \textbf{0.71} & \textbf{6.79} \\ \bottomrule
\end{tabular}%
}
\end{table}

\begin{figure}[H]
\centering
\includegraphics[width=0.6\textwidth]{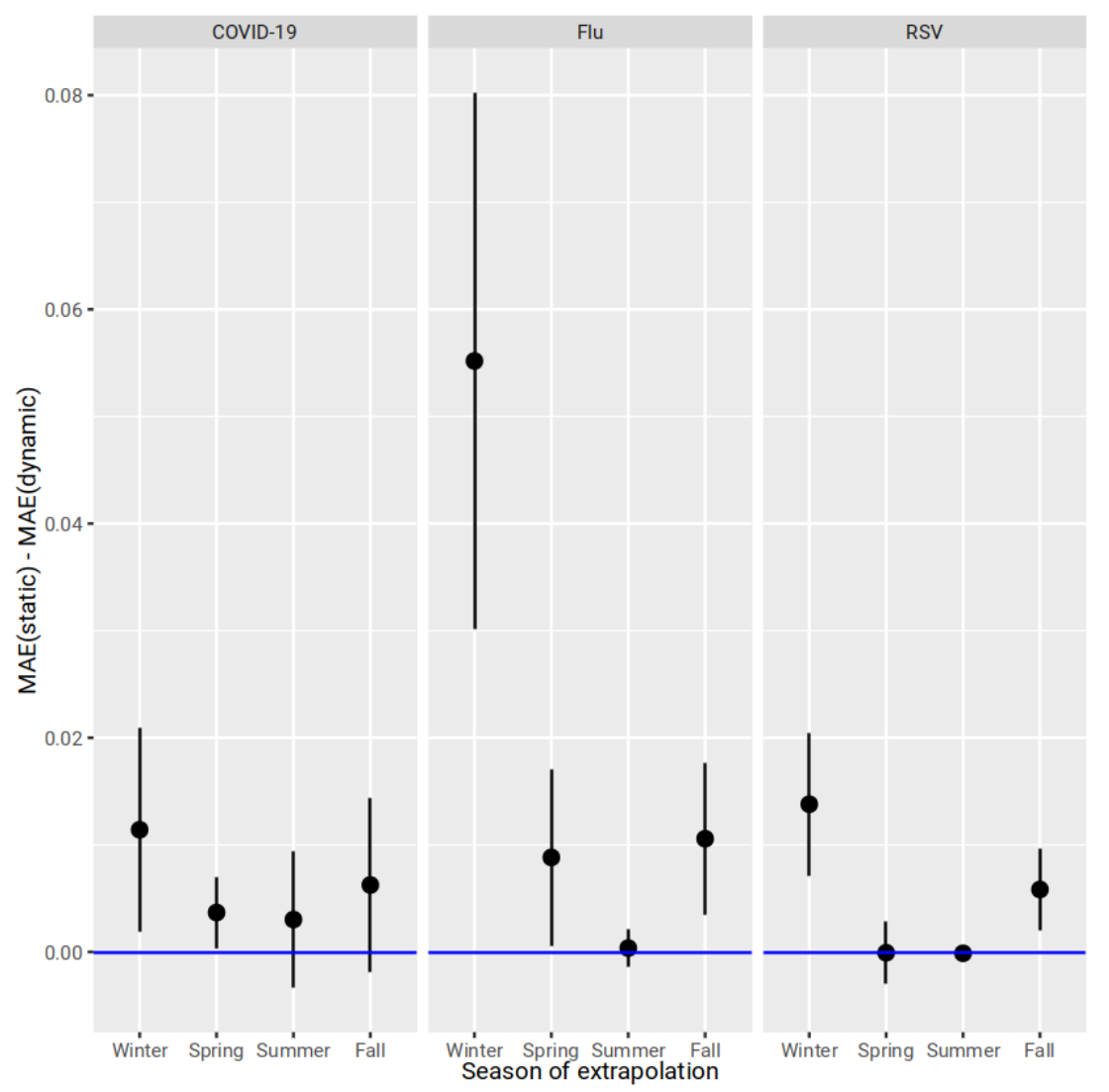} 
\caption{Reduction in Mean Absolute Extrapolation Error by Dynamic vs. Static Population Dynamics Foundations: Comparison across monthly COVID-19, flu, and RSV emergency department visits over a two-year period. The results show that using dynamic (time-matched) PDFM embeddings consistently and significantly improves extrapolation performance compared to using static embeddings trained in July 2023.}
\label{fig:mae_reduction}
\end{figure}

\paragraph{Temporal Population Dynamics Embeddings}
To demonstrate the value of the temporal embeddings, we applied them to the task of per-month extrapolation on the COVID-19, flu, and RSV emergency department visits datasets from PopHIVE \citep{pophive2025}. Similarly to \citep{agarwal2024general}, each month's extrapolation task consists of predicting the percentage of emergency department visits caused by the given condition on all the counties in a fixed holdout set of ten US states. Figure~\ref{fig:mae_reduction} shows the reduction in mean absolute error (MAE) attained by using a time-varying rather than static \textit{Population Dynamics Foundations} embedding. In particular, “dynamic \textit{Population Dynamics Foundations}” uses the current month’s \textit{Population Dynamics Foundations} embedding to do the extrapolation while “static \textit{Population Dynamics Foundations}” uses embeddings trained in July 2023. Each point represents the difference in MAE across all counties in the validation set for all months in the season (across a 24-month period beginning July 2023). Error bars show +/- 2 standard errors in the estimate of the difference in mean absolute error (calculated under the assumption that months and locations are independent). The data suggests temporal embeddings consistently provide an advantage over static embeddings in a number of season-disease pairs. This difference is particularly high in the winter and fall months when the burden of disease is much higher.

\paragraph{External Independent Validation of Population Dynamics Static Embeddings}
To demonstrate the real-world applicability and robustness of our model, the \textit{Population Dynamics Foundations} static embeddings were evaluated by independent third parties. Table~\ref{tab:external_validation} presents the results of these external validations, which span a range of applications and domains, confirming the practical utility and effectiveness beyond controlled academic benchmarks.

{ %
\centering 
\begin{tabularx}{\textwidth}{ 
  >{\hsize=0.9\hsize}X  
  >{\hsize=1.0\hsize}X  
  >{\hsize=0.6\hsize}X  
  >{\hsize=1.5\hsize}X  
  >{\hsize=0.8\hsize}X  
}

\caption{Independent External Validation of Population Dynamics Foundations: We show examples of research partners independently validating the Population Dynamics Foundations in a variety of use cases across commercial and social good target variables and geospatial models including extrapolation and forecasting.}
\label{tab:external_validation} \\
\toprule
Task & Method & Metric & Result & Reference \\ \midrule
\endfirsthead

\toprule
\multicolumn{5}{c}{\tablename~\thetable: Continued} \\
\midrule
Task & Method & Metric & Result & Reference \\ \midrule
\endhead

\midrule
\multicolumn{5}{r}{\textit{Continued on next page}} \\
\endfoot

\bottomrule
\endlastfoot

Interpolate CARTO’s Human Activity Index across the United States & Population Dynamics Foundations embeddings with a Random Forest Regressor & $R^2$ & The model achieved a $R^2$ of 0.882 on the test split, explaining almost 90\% of their proprietary index. & CARTO \citep{Alvarez2025} \\
\midrule
Interpolate Iowa's liquor sales in the United States & Population Dynamics Foundations embeddings and demographics covariates with a Random Forest Regressor & $R^2$ & The ensemble model combining Population Dynamics Foundations and demographics features performed the best, with a $R^2$ of 0.707. Population Dynamics Foundations alone yielded a $R^2$ of 0.666, which is 0.097 higher than demographics alone. & CARTO \citep{Alvarez2025b} \\
\midrule
Interpolation and super-resolution of insurance premium or non-renewal rates in the United States & Population Dynamics Foundations model embeddings with a XGBoost Regressor & Top-2 accuracy & The model achieved 93\% top-2 accuracy in classifying home insurance premiums at the county-level and 89.5\% top-2 accuracy at zip code level. & SustGlobal \citep{ballard2025populous} \\
\midrule
Medium to long-term (1-12 months) forecasting of Dengue in Brazil & Various time-series, foundation, and machine learning models were evaluated with and without inclusion of the Population Dynamics Foundations model embeddings. & Root Mean Square Error (RMSE), Weighted Interval score (WIS), $R^2$ & The model yielding the best fit varied across seasons, with ensemble performance consistently among the best-performing models (e.g., $R^2$ = 0.849 and 0.901 for models without and with the inclusion of Population Dynamics Foundations models, respectively). \newline Population Dynamics Foundations provided greatest benefit when applied with TimesFM, especially at mid-long range forecasting, increasing $R^2$ from 0.500 to 0.686 (6 months) and from 0.456 to 0.656 (12 months). & University of Oxford \newline Kraemer et al. (in preparation) \\
\midrule
Spatiotemporal risk modeling of Poliovirus in Nigeria & Population Dynamics Foundations incorporated via ridge regression, alongside fixed and spatial random effects commonly used risk prediction models & DlC/WAIC & Incorporating Population Dynamics Foundations improved model fit, as measured by DIC and WAIC. Predicted risk differed meaningfully (>2-fold increase or decrease from the base model) in 18\% of local government areas (LGAs) & Institute for Disease Modeling \newline Chabot-Couture, Voorman et al. (preliminary study) \\
\midrule
Super-resolution mapping of Measles in the United States & Population Dynamics Foundations to upscale resolution & Spatial Resolution & Increased spatial resolution of measles vulnerability maps from county-level to ZCTA-level (zip code). & Icahn School of Medicine at Mt. Sinai, Boston Children’s Hospital,Harvard Medical School \newline Zhou, Brownstein, Rader et al. (in preparation) \\ 

\end{tabularx}
} %

\subsubsection{Combining Earth AI Models: Predictive Applications}
With the individual performance of our Imagery and Population models validated, we now address our hypothesis that their combination yields synergistic insights. The following experiments are designed specifically to quantify the performance lift gained by fusing complementary Earth AI models.

\paragraph{Predicting FEMA Risk Scores with Population Dynamics Foundations + AlphaEarth Foundations}
FEMA risk scores \citep{FEMA2025} measure communities’ relative risk in the event of natural disasters, and combine Expected Annual Loss, Social Vulnerability, and Community Resilience. A model that predicts risk scores for natural disasters such as earthquakes, tornadoes, floods, and landslides helps communities better prepare and plan for natural disasters. Risk prediction for these disasters is inherently bimodal, depending on two distinct feature sets: the physical likelihood and severity of the hazard, which can be represented by remote sensing data (e.g., elevation, land cover, proximity to coastlines), and the community's capacity to absorb and recover from a disaster, which is captured by population-level attributes (e.g., density, wealth, age distribution). Therefore, to better predict FEMA risk, we used features derived from two complementary foundation models: \textit{AlphaEarth Foundations} (for landscape features) and \textit{Population Dynamics Foundations} (for socioeconomic features). Both are available for general use, making them ideal for this predictive task.

To test our ability to predict the FEMA risk scores at the US census tract level, we divided the census tracts by county into training and test sets. The census tracts in 20\% of the counties are used as the test set, and the remaining 80\% for training and validation. We derived census tract level features from \textit{Population Dynamics Foundations} and the recently released \textit{AlphaEarth Foundations} \citep{brown2025alphaearth} for the contiguous USA and used these features to train a gradient boosted trees model for each target variable. Our analysis found that combining \textit{Population Dynamics Foundations} with \textit{AlphaEarth} embeddings yielded an average 11\% relative increase in $R^2$ when predicting the 20 FEMA risk score indices (Table~\ref{tab:fema_risk}), compared to models which leveraged \textit{Population Dynamics Embeddings} or \textit{AlphaEarth Foundations} embeddings alone.

\begin{table}[H]
\centering
\caption{$R^2$ metric for predicting 20 FEMA risk scores in census tracts in 20\% of the counties. Models are trained on census tract level labels in 80\% of the counties, using features derived from AlphaEarth Foundations embeddings, Population Dynamics Foundations embeddings, and both sets of embeddings. Models using both sets of embeddings outperformed those using individual feature sets on most tasks and resulted in a 11\% increase in $R^2$ on average. Bold indicates the highest value per task (label).}
\label{tab:fema_risk}
\resizebox{\textwidth}{!}{%
\begin{tabular}{lccc}
\toprule
Target Label & AlphaEarth Foundation & \begin{tabular}[c]{@{}c@{}}Population Dynamics \\ Foundations\end{tabular} & \begin{tabular}[c]{@{}c@{}}AlphaEarth + \\ Population Dynamics\end{tabular} \\ \midrule
Avalanche Risk & 0.13 & 0.11 & \textbf{0.15} \\
Coastal Flooding Risk & 0.33 & 0.24 & \textbf{0.38} \\
Cold Wave Risk & 0.66 & 0.66 & \textbf{0.68} \\
Drought Risk & \textbf{0.67} & 0.50 & \textbf{0.67} \\
Expected Annual Loss Score & 0.59 & 0.55 & \textbf{0.62} \\
Earthquake Risk & 0.72 & 0.84 & \textbf{0.86} \\
Hail Risk & 0.56 & \textbf{0.59} & \textbf{0.59} \\
Hurricane Risk & 0.77 & 0.81 & \textbf{0.83} \\
Heat Wave Risk & 0.58 & 0.58 & \textbf{0.63} \\
Ice Storm Risk & 0.46 & 0.51 & \textbf{0.53} \\
Landslide Risk & 0.47 & 0.53 & \textbf{0.56} \\
Lightning Risk & 0.39 & 0.47 & \textbf{0.49} \\
Resilience Score & 0.65 & 0.71 & \textbf{0.74} \\
Riverine Flooding Risk & 0.26 & 0.26 & \textbf{0.29} \\
Risk Score & 0.53 & 0.49 & \textbf{0.60} \\
Social Vulnerability Score & 0.30 & 0.44 & \textbf{0.48} \\
Strong Wind Risk & 0.66 & 0.71 & \textbf{0.73} \\
Tornado Risk & 0.73 & 0.71 & \textbf{0.75} \\
Wildfire Risk & \textbf{0.77} & 0.54 & 0.76 \\
Winter Weather Risk & 0.60 & 0.54 & \textbf{0.61} \\ \midrule
All metrics (mean) & 0.54 & 0.54 & \textbf{0.60} \\ \bottomrule
\end{tabular}%
}
\end{table}

\paragraph{Predicting Health Statistics with Population Dynamics Foundations + AlphaEarth Foundations}
To test out the embeddings in a different domain, we also applied the same modeling setup described above to predict 21 health indicator variables from the CDC. We found that when we combined both embeddings, it improved the $R^2$ over \textit{Population Dynamics Foundations} alone by 7\% and over \textit{AlphaEarth Foundations} alone by 43\% (Table~\ref{tab:cdc_health}). This set of labels was retrieved from Google Data Commons (see Methods section in Appendix \ref{subsubsec:appendix_population_dynamics_foundations}).

\begin{table}[H]
\centering
\caption{$R^2$ metric for predicting 21 CDC health indicator variables in census tracts in 20\% of the counties. Models are trained on census tract level labels in 80\% of the counties, using features derived from Population Dynamics Foundations embeddings, AlphaEarth Foundations embeddings, and both sets of embeddings. Models using both sets of embeddings outperformed those using individual feature sets on all tasks. Bold indicates the highest value per task (label).}
\label{tab:cdc_health}
\resizebox{\textwidth}{!}{%
\begin{tabular}{lccc}
\toprule
Target Label & AlphaEarth Foundations & \begin{tabular}[c]{@{}c@{}}Population Dynamics \\ Foundations\end{tabular} & \begin{tabular}[c]{@{}c@{}}AlphaEarth + \\ Population Dynamics\end{tabular} \\ \midrule
HighCholesterol & 0.41 & 0.53 & \textbf{0.56} \\
PhysicalHealthNotGood & 0.36 & 0.53 & \textbf{0.57} \\
Stroke & 0.32 & 0.51 & \textbf{0.54} \\
BingeDrinking & 0.31 & 0.49 & \textbf{0.50} \\
PhysicalInactivity & 0.48 & 0.61 & \textbf{0.66} \\
ReceivedAnnualCheckup & 0.62 & 0.68 & \textbf{0.71} \\
Cancer & 0.47 & 0.58 & \textbf{0.64} \\
Diabetes & 0.39 & 0.59 & \textbf{0.61} \\
MentalHealthNotGood & 0.38 & 0.49 & \textbf{0.57} \\
CoronaryHeartDisease & 0.40 & 0.56 & \textbf{0.58} \\
HighBloodPressure & 0.45 & 0.61 & \textbf{0.62} \\
CholesterolScreening & 0.42 & 0.52 & \textbf{0.59} \\
DentalVisit & 0.53 & 0.63 & \textbf{0.69} \\
Asthma & 0.28 & 0.48 & \textbf{0.53} \\
ChronicKidneyDisease & 0.25 & 0.43 & \textbf{0.46} \\
Arthritis & 0.46 & 0.62 & \textbf{0.65} \\
COPD & 0.47 & 0.62 & \textbf{0.64} \\
BPMed & 0.39 & 0.54 & \textbf{0.59} \\
Obesity & 0.41 & 0.61 & \textbf{0.65} \\
SleepLessThan7Hours & 0.49 & 0.61 & \textbf{0.65} \\
Smoking & 0.46 & 0.57 & \textbf{0.62} \\ \midrule
All metrics (mean) & 0.42 & 0.56 & \textbf{0.60} \\ \bottomrule
\end{tabular}%
}
\end{table}

\paragraph{Predicting Disaster Damage with Cyclones Forecast + Population Dynamics Foundations}
When a natural disaster hits, getting resources and support to affected people, businesses and cities as fast as possible is crucial. Hurricanes consistently rank amongst the most damaging natural disasters in the USA, with wind-related damage accounting for the majority of losses. A model capable of predicting vulnerable buildings that allows reliable risk assessment can accelerate disaster relief to affected businesses and homeowners.

\begin{figure}[H]
\centering
\includegraphics[width=\textwidth]{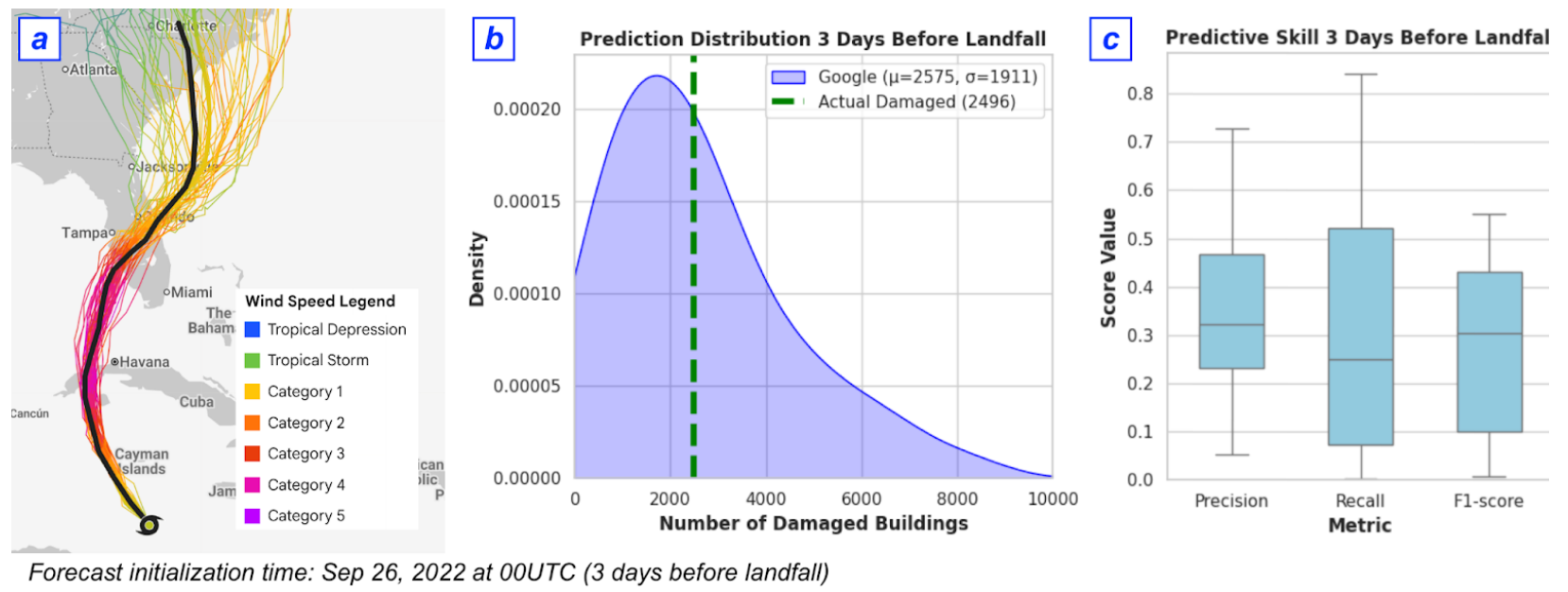}
\caption{Cyclone damage forecasting when adding Population Dynamics Foundations: a) Hurricane track and wind speed forecasts by Google’s Experimental Cyclone Forecast for Hurricane Ian at the initialization time of Sep 26, 2022 at 00UTC (3 days before landfall). The observed hurricane track is shown in bold black line. b) Distribution of the predicted number of damaged buildings using Bellwether’s model. Actual number of damaged buildings (ground truth) is shown with a green dashed line. c) Boxplots illustrating the performance of Bellwether’s predictive model. Note that the metrics (Precision, Recall and F1 for the damaged buildings class) only focus on a single snapshot in time and should not be generalized to other events or lead times.}
\label{fig:cyclone_damage}
\end{figure}

Bellwether\footnote{\url{https://x.company/projects/bellwether/}} has utilized hurricane tracks and wind speed forecasts of Google’s experimental cyclone model, insights from local urban characteristics (e.g. building heights from Google’s Open Buildings dataset) and \textit{Population Dynamics Foundations} embeddings to train a model and predict which buildings will incur wind damage. The model was tested on Hurricane Ian three days before landfall, and the forecasted tracks and the associated wind speeds are shown in Figure~\ref{fig:cyclone_damage}a.

Focusing on Hurricane Ian’s damage forecasts three days before landfall, our combined model shows strong performance. The expected number of damaged buildings is of high importance to determine the magnitude of relief funds. Figure~\ref{fig:cyclone_damage}b shows the distribution of the forecasted number of damaged buildings (among 50 ensemble members). In this dataset, 2,496 buildings were ultimately damaged. Three days before landfall, the model predicted an average of 2,575 damaged buildings—an error of only 3\% from the observed number. Figure~\ref{fig:cyclone_damage}c displays the distribution of performance metrics for the damaged buildings class across the 50 ensemble members.

We note that this is a single retrospective analysis; however, it serves as a powerful demonstration of the model’s potential in data-sparse, high-urgency scenarios.

\paragraph{Forecasting Cholera Risk with TimesFM + Population Dynamics Foundations + Weather Forecasting}
The Democratic Republic of Congo (DRC) faces a persistent public health crisis from endemic cholera. In collaboration with WHO AFRO, utilizing data from WHO's Integrated Disease surveillance team we evaluate a sub-national forecasting model for cholera cases in the DRC. The forecasting model utilized \textit{TimesFM 2.0}, Google's large foundation model for time series forecasting, \textit{Population Dynamics Foundations} embeddings as static numerical covariates for each province, and dynamic covariates with weather forecasting of meteorological variables, specifically precipitation and temperature, which are known drivers of cholera transmission.

An initial zero-shot forecast was generated using only historical weekly cholera case counts for 24 DRC provinces. This ``out-of-the-box'' application of \textit{TimesFM} established a strong performance baseline, demonstrating significantly higher accuracy than standard forecasting models like SARIMAX and Prophet. At a national level, for a 20-week forecast horizon, the baseline \textit{TimesFM} model achieved a 35.4\% RMSE improvement over Prophet. Furthermore, augmenting the baseline model with dynamic weather forecasts resulted in a performance improvement of 32.7\%. At a regional level, incorporating \textit{Population Dynamics Foundations} embeddings as covariates reduced the forecast RMSE by 5\% when compared to baseline \textit{TimesFM}. These results confirm the model's effectiveness in capturing trends and seasonality in the raw data, especially when combined with Earth AI models.

\subsection{Orchestration: Solving Complex Queries with the Geospatial Reasoning Agent}
The preceding results demonstrate the results of our individual models and power of combining models for well-defined predictive tasks. However, real-world challenges often require complex, multi-step reasoning. We now evaluate the final component of our system: the Geospatial Reasoning Agent - designed to orchestrate all of our models and tools to answer such complex queries, as described earlier.

To evaluate our Geospatial Reasoning Agent’s capabilities, we developed a new evaluation-set featuring questions that test an agent's ability to use the tools across different domains and reason correctly about queries that require descriptive or data retrieval along with queries that are more analytical or relational in nature. For each prompt in our Q\&A evaluation-set, we had a specific quantitative or list-based ground truth answer to evaluate performance.

To assess problem-solving scenarios that integrate a predictive model, we constructed a Crisis Response evaluation-set that focuses on queries related to Crisis Response with a corresponding rubric to evaluate each. Each of these queries combines a predictive model (RemoteSensing/Population Dynamics Foundations or flood/weather/cyclone forecasts) and a component related to retrieval and relational by combining multiple tools.

For both benchmarks, we developed custom autoraters (see Appendix \ref{subsubsec:appendix_benchmark_generation_and_eval_metrics} for details) that focused on either evaluating the ground truth answer or comparing the response to a detailed rubric.

\subsubsection{Q\&A evaluation-set: Fact-finding and Data Analytics}
To assess the agent's ability to retrieve and reason over data, we designed a Q\&A evaluation-set with 100 questions and manually verified answers spanning Places, People, Weather, and the Environment. The questions test both direct data retrieval and more complex analytical reasoning that requires relating multiple variables or time periods. All questions included in this evaluation-set can be answered through publicly available datasets and APIs. The questions can also be divided into two semantic categories:
\begin{itemize}
    \item \textbf{Descriptive and Retrieval.} We had two main design criteria for this category of questions. The first is to define unambiguous questions with verifiable answers. The second is that the answers need to be time-invariant in the short term. An example prompt from this category is “Respond by filling in the blanks: There are \_\_ botanical gardens in Lincoln, Nebraska”.
    \item \textbf{Analytical and Relational.} These questions are more complex than the ones in the previous category. In addition to satisfying the criteria above, their answers must involve additional analysis to relate multiple places, sociodemographic variables, or time periods. For example, ranking places, describing changes over time, or computing correlation coefficients. An example from this category is “Which county with a population of at least 3,000 had the highest number of PhDs per capita in 2022? Respond by filling in the blanks: County: \_\_, PhDs per capita: \_\_”.
\end{itemize}

All questions in the evaluation set are fill-in-the-blanks. Scores between 0.0 and 1.0 (higher is better) are computed by comparing the values that the agent filled in against the values in the gold answer, while ignoring all surrounding text. Textual values are scored using ROUGE-L F1 and numerical values are scored based on a clamped absolute percentage error. See Appendix \ref{para:fact_finding} for a detailed description of the scoring method, as well as sample prompts and responses from the Geospatial Reasoning Agent.

The performance of our Geospatial Reasoning Agent was compared against baseline Gemini 2.5 Pro and Gemini 2.5 Flash agents with no custom tools. The baseline agents are also implemented in ADK, with Gemini's built-in Google Search, Grounding with Google Maps, and Code Execution tools enabled (details in Appendix \ref{subsubsec:appendix_baseline_gemini_configuration}), giving them the potential to answer the benchmark questions from parametric knowledge, online information retrieval, and code-based analysis. As the results in Table~\ref{tab:qa_results} indicate, the specialized Geospatial Reasoning Agent outperforms the baselines across all domains of this evaluation-set, achieving an overall score of $0.82 \pm 0.02$ compared to $0.50 \pm 0.01$ (Pro) and $0.39 \pm 0.02$ (Flash). It outperformed the baseline Gemini 2.5 Pro agent by 37\% in the Descriptive and Retrieval category, and 124\% in the more complex Analytical and Relational category, for an overall 64\% higher score. A similar increase in the performance gap can also be observed between the baseline Pro and Flash agents when moving from the Descriptive to the Analytical category.

\begin{table}[H]
\centering
\caption{Performance metrics for the Q\&A evaluation-set, comparing the Earth AI Geospatial Reasoning Agent's performance against baseline agents without custom tools: The number of questions in each domain are included in brackets, and for each domain the questions are evenly split across the Descriptive and Analytical categories. The scores are averaged across 5 runs, with $\pm$ values indicating the 95\% confidence intervals across runs. Bold indicates the highest value per domain.}
\label{tab:qa_results}
\resizebox{\textwidth}{!}{%
\begin{tabular}{llccccc}
\toprule
& & \multicolumn{4}{c}{Domain (number of questions)} & \\
\cmidrule(lr){3-6}
Category & Agent & Places (30) & \begin{tabular}[c]{@{}c@{}}People \& \\ Communities (40)\end{tabular} & \begin{tabular}[c]{@{}c@{}}Weather \& \\ Environment (20)\end{tabular} & \begin{tabular}[c]{@{}c@{}}Multiple \\ Domains (10)\end{tabular} & Overall (100) \\ \midrule
\multirow{3}{*}{\begin{tabular}[c]{@{}l@{}}Descriptive and \\ Retrieval (50)\end{tabular}} & Gemini 2.5 Flash & $0.41 \pm 0.03$ & $0.81 \pm 0.01$ & $0.54 \pm 0.05$ & $0.47 \pm 0.03$ & $0.60 \pm 0.01$ \\
 & Gemini 2.5 Pro & $0.44 \pm 0.04$ & $0.86 \pm 0.02$ & $0.72 \pm 0.01$ & $0.53 \pm 0.06$ & $0.67 \pm 0.01$ \\
 & Geospatial Reasoning & \textbf{0.88 $\pm$ 0.05} & \textbf{0.97 $\pm$ 0.03} & \textbf{0.78 $\pm$ 0.04} & \textbf{1.00 $\pm$ 0.00} & \textbf{0.91 $\pm$ 0.01} \\ \midrule
\multirow{3}{*}{\begin{tabular}[c]{@{}l@{}}Analytical and \\ Relational (50)\end{tabular}} & Gemini 2.5 Flash & $0.29 \pm 0.07$ & $0.11 \pm 0.05$ & $0.15 \pm 0.04$ & $0.10 \pm 0.03$ & $0.17 \pm 0.04$ \\
 & Gemini 2.5 Pro & $0.46 \pm 0.04$ & $0.23 \pm 0.02$ & $0.37 \pm 0.07$ & $0.22 \pm 0.08$ & $0.33 \pm 0.03$ \\
 & Geospatial Reasoning & \textbf{0.57 $\pm$ 0.04} & \textbf{0.86 $\pm$ 0.06} & \textbf{0.64 $\pm$ 0.14} & \textbf{0.95 $\pm$ 0.08} & \textbf{0.74 $\pm$ 0.04} \\ \midrule
\multirow{3}{*}{Overall (100)} & Gemini 2.5 Flash & $0.35 \pm 0.05$ & $0.46 \pm 0.03$ & $0.35 \pm 0.03$ & $0.28 \pm 0.03$ & $0.39 \pm 0.02$ \\
 & Gemini 2.5 Pro & $0.45 \pm 0.01$ & $0.54 \pm 0.02$ & $0.54 \pm 0.04$ & $0.38 \pm 0.03$ & $0.50 \pm 0.01$ \\
 & Geospatial Reasoning & \textbf{0.73 $\pm$ 0.03} & \textbf{0.91 $\pm$ 0.02} & \textbf{0.71 $\pm$ 0.08} & \textbf{0.98 $\pm$ 0.04} & \textbf{0.82 $\pm$ 0.02} \\ \bottomrule
\end{tabular}%
}
\end{table}

\subsubsection{Geospatial Reasoning in Action: Crisis Response Case Study}
In this section, we evaluate the agent's ability to select and utilize the appropriate foundation models and as well as expert agents for complex queries. While our case-study primarily focuses on the crisis-response use-case, the underlying capabilities—such as data gathering, correctly executing inference functions, and dynamic model training—are domain-agnostic.

Our evaluation-suite consists of 10 constrained prompts that integrate capabilities such as flood, cyclone, and weather forecasting, as well as Remote Sensing, TimesFM, and Population Dynamics Foundations, alongside geospatial and demographics sub-agents. Performance was measured against a list of rubrics accompanying each prompt, rather than a final “gold answer”. This approach allowed the evaluation to focus on the logical soundness of the agent's reasoning path instead of the performance of each of the predictive models which have been captured above.

To capture response variability, we executed each prompt 10 times for both our agent and the same baseline Gemini 2.5 Pro agent used in the Q\&A benchmark. The mean Likert scores from these iterations $\pm$ 95\% confidence intervals, detailed in Table~\ref{tab:crisis_response}, reveal that our agent consistently achieved superior performance in these complex scenarios. This result underscores the distinct advantage of providing a model with direct access to specialized geospatial data, specialized tools and models, and custom reasoning.

\begin{table}[H]
\centering
\caption{Crisis Response Case Study: An overview of our benchmark results including the crisis type, capabilities tested and average $\pm$ 95\% confidence intervals of the Likert scores across 10 runs per prompt comparing Gemini 2.5 Pro with our Geospatial Reasoning Agent for the 10 prompts that make up our Crisis Response Case Study Benchmark. Bold indicates the higher value for each task (crisis).}
\label{tab:crisis_response}
\resizebox{\textwidth}{!}{%
\begin{tabular}{llp{5.5cm}cc}
\toprule
ID & Crisis & Capabilities Tested & \begin{tabular}[c]{@{}c@{}}Gemini 2.5 Pro \\ Likert Score\end{tabular} & \begin{tabular}[c]{@{}c@{}}Geospatial Reasoning Agent \\ Likert Score\end{tabular} \\ \midrule
1 & Flood & Floods Forecast + Places & $0.18 \pm 0.21$ & \textbf{0.80 $\pm$ 0.30} \\
2 & Flood & Floods Forecast + Demographics & $0.28 \pm 0.21$ & \textbf{0.80 $\pm$ 0.30} \\
3 & Typhoon & Cyclones Forecast + Location & $0.23 \pm 0.21$ & \textbf{0.93 $\pm$ 0.12} \\
4 & Hurricane & Cyclones Forecast + Demographics & $0.45 \pm 0.14$ & \textbf{0.80 $\pm$ 0.18} \\
5 & Drought & Weather Forecast + Earth Engine & $0.58 \pm 0.09$ & \textbf{0.88 $\pm$ 0.09} \\
6 & Extreme Heat & Weather Forecast + Demographics & $0.76 \pm 0.12$ & \textbf{1.00 $\pm$ 0.00} \\
7 & Fire & RemoteSensing Foundation + Places & $0.40 \pm 0.15$ & \textbf{0.84 $\pm$ 0.08} \\
8 & Flood & RemoteSensing Foundation + Demographics & $0.35 \pm 0.21$ & \textbf{0.88 $\pm$ 0.09} \\
9 & Hurricane & TimesFM + Demographics & $0.00 \pm 0.00$* & \textbf{1.00 $\pm$ 0.00} \\
10 & Hurricane & PopulationDynamics Foundations + Demographics & $0.25 \pm 0.17$ & \textbf{0.91 $\pm$ 0.09} \\ \midrule
\multicolumn{3}{l}{Mean Across Prompts (excluding row 9*)} & $0.38 \pm 0.17$ & \textbf{0.87 $\pm$ 0.14} \\ \bottomrule
\end{tabular}%
}
\captionsetup{labelsep=none}
\caption*{\small{* Gemini was consistently unable to find authoritative labor statistics for 2024, so it’s unable to meaningfully answer the question.}}
\end{table}

\begin{figure}[H]
\centering
\includegraphics[width=\textwidth]{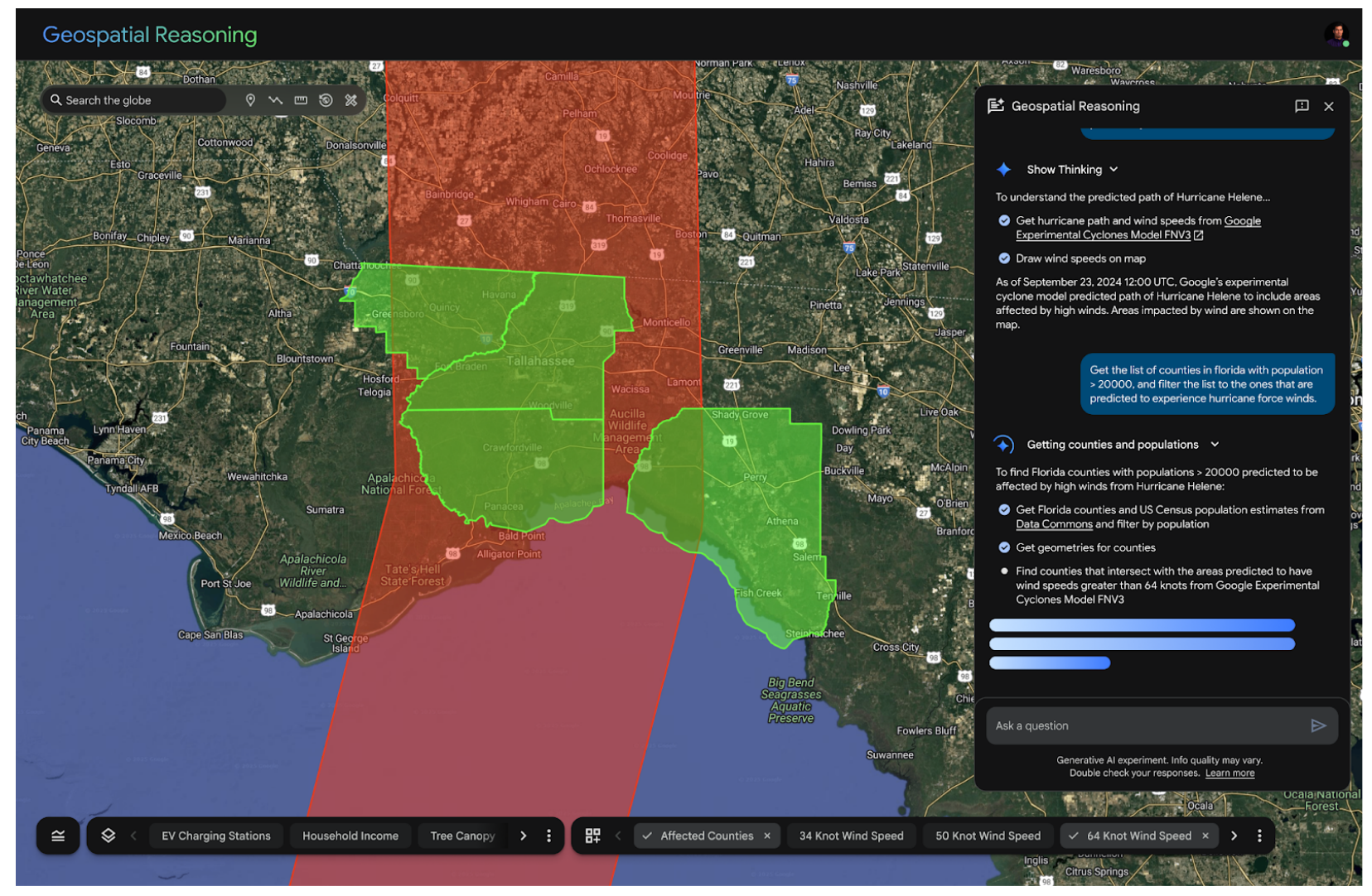}
\caption{A demonstration of the Earth AI Geospatial Reasoning Agent in action. The agent is first given the task to find counties in Florida that are predicted to experience hurricane-force winds by Google's experimental Cyclone model. For this It fetches the hurricane forecast and extracts the area of hurricane-force winds, highlighted in red. The agent also fetches the list of counties in Florida and their population from the U.S. Census Bureau via Data Commons, filtering for counties with a population over 20,000. The agent then looks at the spatial intersection of these counties and the hurricane-force wind polygon, and adds the resulting four counties on the map in green.}
\label{fig:agent_demo}
\end{figure}

To illustrate this approach, Figure~\ref{fig:agent_demo} shows an example prompt similar to ID 4 from our crisis response benchmark related to cyclone forecast and its implications for at-risk communities. The exact prompt is “As of September 23, 2024 12:00 UTC, what did Google's cyclones model say the predicted path of Hurricane Helene was? Get the list of counties in florida with population > 20000, and filter the list to the ones that are predicted to experience hurricane force winds.”, where we have split up the questions to showcase the interactive flow between a user and our agent. The accompanying response shows how our agent leverages both the experimental cyclone model and Google Data Commons to address the prompt. For the complete list of prompts and rubrics, along with an example Agent and Gemini response, see Appendix \ref{para:crisis_response_case_study} “Crisis Response Case Study”.

\section{Discussion}
Our work introduces and validates a suite of specialized foundation models—\textit{RemoteSensing Foundations} for high-resolution imagery, \textit{Population Dynamics Foundations} for people and places, and advanced models for environmental prediction—that serve as the core modalities of our integrated Earth AI framework. The findings indicate a path forward for geospatial AI that moves beyond siloed, task-specific models toward integrated ecosystems. The results strongly suggest that the future of the field lies in synergistic interplay between specialized models—orchestrated by an advanced Gemini-based reasoning engine to unlock a more comprehensive and actionable understanding of our planet. This framework is also designed to lower the barrier to entry, enabling domain experts to conduct sophisticated spatial analysis without deep expertise in data science or GIS by automating the discovery and integration of relevant models and datasets.
\subsection{Remote Sensing Foundations}
We have shown that by creating specialized novel datasets and applying recent AI architectures and training methods, we are able to bridge the gap between everyday computer vision capabilities and the distinct demands of earth observation applications leading to several notable accomplishments. First, we demonstrated the efficacy of VLMs trained with our novel datasets, achieving state-of-the-art performance on zero-shot cross-modal retrieval benchmarks. We then showed how the specialized VLM-derived representations enable the OVD model to extend to free-text categories targeting the remote sensing domain, and to enhance performance and address vocabulary mismatch challenges by using a novel Few-shot algorithm. Finally, our self-supervised pre-training method for backbone models demonstrated robust generalization capacity across diverse downstream tasks.

\paragraph{Limitations and Future Work:} Key limitations include the current focus on high-resolution RGB imagery and an evaluation that does not yet encompass temporal downstream tasks. Future work will focus on expanding these models to support more zero-shot natural language capabilities, temporal tasks and a broader range of remote sensing sensors and imagery configurations, such as oblique, multispectral and hyperspectral imagery.

\subsection{Population Dynamics Foundations}
\textit{Population Dynamics Foundations} provides an accessible yet powerful lens into the complex dynamics of human populations, a critical component for any human-centric analysis. The model's value has been demonstrated through successful external independent validation across multiple applications, confirming its real-world utility. Furthermore, its recent expansion to global coverage and the introduction of temporal embeddings represent a significant leap in its analytical power.

\paragraph{Limitations and Future Work:} Future work will focus on three key areas. First, we aim to extend the temporal look-back period beyond the current few months to better model long-term seasonal or multi-year trends. Second, we plan to incorporate a wider array of privacy-preserving population signals and push towards finer geographic and temporal granularities. Finally, our interpolation evaluations, while promising, will be expanded to test against systematic data gaps that better reflect real-world scenarios where outages may not be random.

\subsection{Combining Earth AI Models: Predictive Applications}
The consistent outperformance of combined core Earth AI models is a direct reflection of the complementary and synergistic nature of these core Earth AI models. This enabled significant accuracy gains seen in our cholera forecasting, FEMA risk prediction, and hurricane damage assessments. By integrating diverse data modalities, the system develops a more holistic and robust understanding of complex real-world phenomena than any single model could achieve on its own. This successful integration validates that a compositional, multi-modal approach is a critical step toward unlocking a deeper understanding of our planet.

\paragraph{Limitations and Future Work:} While this paper highlights the effectiveness of combining specialized models, a significant future advancement would be to develop a single, unified meta-Earth model to replace the current modality-specific foundation models. Such a model, trained concurrently on imagery, environmental data, and population signals, could create a shared, multi-modal representation of the planet, leading to a more holistic understanding of the Earth’s systems. The primary technical hurdle we would like to overcome in future work is the alignment of models operating at different spatial and temporal granularities. For instance, our method for aggregating AlphaEarth's 10-meter embeddings to the census tract level can be improved. Furthermore, the combined model currently requires tuning for each new predictive variable, pointing to an opportunity for more generalized fusion methods.

\subsection{Geospatial Reasoning across Earth AI Models}
Our agentic approach represents a significant step forward in making complex geospatial analysis more accessible and scalable. It emulates the workflow of a human analyst—but at a far greater speed and scale—by efficiently querying vast planetary datasets and performing on-the-fly analytics and modeling. The Geospatial Reasoning agent coordinates multiple specialized models into a unified system, and as our crisis-response benchmark shows, can deconstruct complex problems, delegate tasks to specialized tools, and synthesize the intermediate results into coherent and insightful results. This capability creates new opportunities for both specialists and non-specialists to engage with complex geospatial data. Notably, our agent is based on Gemini, so has the same underlying reasoning capabilities but is specifically optimized for geospatial analytics through the agent design, access to geospatial tools and the prompting strategy. We attribute additional improvements in performance in part to the agent having access to highly capable prediction models and task-specific datasets that aren’t available to base LLMs, and partly due to mitigating variability caused by non-specialised data discovery and retrieval.

\paragraph{Limitations and Future Work:} It is important to acknowledge that agentic geospatial reasoning is a rapidly evolving field. While the suite of tools and models presented here marks substantial progress, there remains a vast landscape of additional capabilities to explore and we plan to continue expanding the agent’s domain coverage and reasoning abilities in future work. While the results of both the Q\&A and the Crisis Response evaluation-sets are encouraging, we note that these domain-specific questions are focused on the agent's designed capabilities. This evaluation primarily validates our approach for common retrieval and analysis tasks but does not yet sufficiently test the agent's performance on out-of-distribution queries or in scenarios where specialized tool orchestration fails. Furthermore, while the crisis response case study is illustrative, a more comprehensive evaluation is needed due to the limitations of a rubrics-only based approach (Appendix \ref{para:limitations_with_rubrics}). Crucially though, this work provides a foundation on which to explore the limits of domain generalization and ultimately make advances in the reliability of real-world problem-solving. In future work, we plan to develop a more robust evaluation framework featuring a broader range of tasks, improved rubrics, and human expert review.

\section{Conclusion}
In this work, we introduced Earth AI, a comprehensive approach designed to address the long-standing challenge of analyzing complex and siloed geospatial data. We showcase through rigorous evaluations state-of-the-art foundation models for Imagery, Population, and the Environment with a Gemini-powered reasoning engine. Their true power of the models is unlocked through synergy, where combining models yields significant accuracy gains in complex predictive tasks like disaster relief and public health forecasting.

The agentic reasoning layer successfully automates complex, multi-step analysis, making sophisticated geospatial intelligence more accessible to non-expert users. The findings presented here strongly suggest that the future of the field lies not in siloed models, but in integrated, multi-modal ecosystems orchestrated by advanced AI.

\section*{Acknowledgement}
On behalf of the authors, we would like to express our heartfelt gratitude to Yun Liu, Alex Boulouvalas, Aliza Hoffman, David Fork, David Menasche, Deborah Cohen, Dong Sun, Ferran Alet, Gabriela Adler, Gila Loike, Jeffrey Cardille, Jeremy Amez-Droz, Joel Conkling, Krishna Sapkota, Lambert Rosique, Peter Battaglia, Prem Ramaswami, Roni Chernyak, Sam Roy, Yichen Zhou, Wojciech Sirko and the AlphaEvolve team for their input to the manuscript.

\bibliography{main}

\appendix
\section{Detailed Methods}
\label{sec:appendix_detailed_methods}
\subsection{Earth AI Models}
\label{subsec:appendix_earth_ai_models}

\subsubsection{Remote Sensing Foundations}
\label{subsubsec:appendix_remote_sensing_foundations}

\paragraph{Vision Language Model (VLM)}
Foundation models in remote sensing (RS) have largely adapted the CLIP architecture. One branch focuses on geographic context, with GeoCLIP \citep{cepeda2024geoclip} based on geolocated personal photography and SatCLIP \citep{klemmer2025satclip} aligning satellite imagery with GPS coordinates. Other methods bridge the cross-view gap: SenClip \citep{jain2025senclip} and Sat2Cap \citep{dhakal2024sat2cap} link overhead imagery to corresponding ground-level photos and their semantics.

Another larger branch trains on explicit image-text pairs which includes RSClip \citep{li2023rsclip}, RemoteClip \citep{liu2024remoteclip}, GeoRSclip \citep{zhang2024georsclip} and RSTransCLIP \citep{elkhoury2024rstransclip}. We continue this line of works and introduce a method for creating a unique remote sensing vision-language dataset to address the absence of rich textual descriptions in remote sensing (RS) images and the scarcity of labeled data. This dataset is then used to train a compact yet robust VLM for remote sensing, designed with the potential to run efficiently on resource-constrained devices using a 400M parameter image encoder. We leverage two pre-trained foundation VLMs, MaMMUT \citep{kuo2023mammut} and SigLIP2 \citep{tschannen2025siglip} with text encoders of 400M parameters.

Our training dataset is composed of three parts: RS-Landmarks, RS-WebLI and RS-Global. The RS-Landmarks dataset consists of 18 million images with high-quality textual descriptions generated with Gemini 1.5 Pro. Initially, the satellite and aerial images are aligned with locations and footprints of places and landmarks extracted from Google Maps, found within the associated images. We feed this information, alongside with the image, to the Gemini model, and use a tailored, curated prompt to generate concise captions for each image. This process yields high-quality and informative captions, describing a diverse set of object categories.

The RS-WebLI dataset consists of aerial and satellite imagery taken from the WebLI dataset \citep{chen2022pali}. We generated RS-WebLI by training aerial and overhead classifiers and thereafter using the classifiers to filter the WebLI dataset. First, we manually classified several hundred RS images in the WebLI dataset, using a simple caption heuristic and manual inspection of the dataset. Second, using the manually labeled sample, we trained an image classifier for remote sensing images and generated a dataset consisting of 40K images. Third, we harnessed the capabilities of crowd computing and initiated a labeling task wherein participants classified images as overhead aerial or satellite imagery, angled aerial imagery, or none of the above. The results were combined with random negatives, yielding a 60K labeled dataset. Lastly, using this larger dataset, we trained new aerial and overhead classifiers and applied them across the entirety of WebLI, filtering the data of which we chose 3 million clean images, creating the RS-WebLI dataset. We then applied the same method on the new 100 billion images WebLI dataset \citep{wang2025scaling} to generate a significantly larger variant of RS-WebLI.

The RS-Global dataset consists of 30M aerial and satellite images collected in the native sensor resolution, varying from 10cm to 10m. The data covers all the land area on earth, excluding the poles and remote islands. It is biased to favour human activity, such as urban and agricultural areas. The imagery was collected from 2003 until 2022, with more weights towards newer images. Each image was then enriched with high-quality textual descriptions generated with Gemini 1.5 Pro. The RS-Global dataset complements the other datasets with a regionally diverse and balanced imagery.

The RS-Landmarks and smaller RS-WebLI datasets were used to fine-tune the RS-MaMMUT model. The larger RS-WebLI variant, RS-Landmarks and RS-Global were used to train RS-SigLIP2.

\paragraph{Open-Vocabulary Object Detection (OVD)}
Recent advancements in large-scale vision-language models have catalyzed a paradigm shift in computer vision, giving rise to Open-Vocabulary Object Detection (OVD). This approach, which allows detectors to identify objects described by arbitrary natural language text, offers unprecedented flexibility compared to traditional detectors limited to predefined categories. This is particularly transformative for remote sensing (RS), where cataloging every possible class is intractable.

Our OVD architecture is based on the OWL-ViT v2 \citep{minderer2023scaling} model, which represents the state-of-the-art in this field. A key advantage of this model is its use of two independent transformers for text and image processing. This architectural design allows for the pre-caching of image embeddings, which enables an efficient retrieval setup.

The training dataset is composed of two parts. The RS-WebLI is enriched with pseudo-annotated bounding boxes generated using an object detection model, OWL-ViT, which identifies objects based on either the image's alt text or a predefined set of labels. Only bounding boxes exceeding a certain confidence threshold are included. The second dataset is a Google internal object detection dataset, which includes a collection of 67,000 aerial images annotated for remote sensing object detection. The dataset contains over 3.5 million labeled instances across 34 object categories.

Our training strategy utilizes the pre-trained OWL-ViT v2 checkpoint. We apply a ``cooldown'' fine-tuning protocol characterized by a linear decay of the learning rate. The model is first fine-tuned on the RS-WebLI dataset for 2 epochs. Following this, the fine-tuning continues for 32 additional epochs on the Google internal object detection dataset, using the same learning rate cooldown schedule.

\paragraph{Few-shot Object Detection}
The Few-shot Retrieval in remote sensing images can significantly improve the performance and address the limitations of zero-shot object detection models when dealing with the ambiguity of natural language queries. The core idea is a cascaded architecture that combines a powerful pre-trained OVD model with a lightweight few-shot classifier.

We use FLAME \citep{refael2025ontheflyovdadaptationflame}, a one-step active learning strategy designed for efficient, real-time adaptation of a zero-shot OVD model to new target classes with minimal human input. This approach prioritizes selecting the most informative samples for training.

First, Uncertainty-based Filtering identifies ambiguous samples by analyzing the density distribution of object embeddings. These embeddings are created by combining image features with a user's text query, and their dimensionality is reduced for efficient analysis. Samples located on the low-density `margins' of the distribution's core are identified as uncertain candidates.

Second, Diversity Sampling refines this selection. It uses k-means clustering to group the uncertain candidates and then selects the most representative sample—the one closest to its cluster's center—from each group. This ensures the final, small subset of samples for user annotation is not only uncertain but also diverse and highly informative.

\paragraph{Pre-training Backbone Foundation Model}
While many remote sensing foundation models focus on vision-language alignment, a distinct class of models emphasizes self-supervised learning directly from visual data. SatMAE \citep{cong2022satmae} and Prithvi \citep{jakubik2024prithvi} focus on pretraining a model specialized on Sentinel-2 satellite imagery, TerraMind  \citep{rupprecht2025terramind} focus on multimodality, etc.

We continue this line of work and present a recipe for pre-training vision only foundation models for remote sensing utilizing self-supervised employing reconstruction-based self-supervised learning methods.

We use an image-only variant of the RS-Global dataset, while scaling it from 30M to 300M images. The images are randomly cropped and resized during training, resulting in a continuous variable resolution, from 10cm up to 10m.

To develop a robust remote sensing foundation model backbone, we adopt a two-stage training strategy. In the first stage, we employ Masked Autoencoder (MAE) pretraining \citep{he2022masked} on the 300M images dataset. This self-supervised learning method forces the encoder to reconstruct masked image patches, enabling the model to capture low-level spectral patterns as well as higher-level spatial structures unique to remote sensing data. While MAE provides strong general-purpose representations, it is limited in task-specific adaptability. To address this, we introduce a second stage of multi-task pretraining (MTP) that jointly optimizes three supervised objectives: scene classification, semantic segmentation, and object detection. This stage enriches the encoder with task-aware features and enhances its robustness to diverse downstream applications. We implement MTP using Alternating Gradient Descent \citep{akbari2023alternating}, where the training alternates between tasks rather than combining them in a single gradient step. This strategy mitigates task interference, allowing different task experts to specialize while still contributing to a unified representation. As a result, the backbone learns features that are more robust, generalizable, and transferable to both seen and unseen downstream tasks across the RS domain. For the MTP we employ Google internal labeled datasets for classification, segmentation and detection.

\subsubsection{Population Dynamics Foundations}
\label{subsubsec:appendix_population_dynamics_foundations}
The Population Dynamics Foundations model is constructed as described in \citep{agarwal2024general} with two notable differences. First, for the global version of the embeddings, when aggregating the Search Trends portion of the embedding, instead of using the counts of the top queries, we now use the counts of the top Knowledge Graph \citep{singhal2012introducing} entities that appear in the Search Trends. For example, the queries \textit{``taylor swift boyfriend''} and \textit{``kc tight end''} would both map to the Knowledge Graph entity \textit{``Travis Kelce''}. Performing this mapping increases the ability of the Search Trends signal to capture patterns across languages and countries and has the added benefit of enhancing privacy. This further enhances the privacy guarantees of the embeddings by mapping many more trends to broader categories and it enables us to connect these representations across Search Trends that might differ in language or more subtle country specific characteristics.

This aggregation was done using search data from October 2023. For each postal code (or similarly sized region in countries where postal boundaries are not available), we computed the frequencies of the top 500 Knowledge Graph entities that were searched at least 20 times. We then ranked all the obtained entities by their global popularity by counting the total number of regions they appeared in across 17 countries: Australia, Belgium, Brazil, Canada, France, Germany, India, Italy, Japan, Mexico, Netherlands, Nigeria, Portugal, Spain, Switzerland, United Kingdom, United States. The top 1000 most popular entities are retained as the feature vocabulary for our model. The least popular entity in the vocabulary appeared in searches in over 9,000 regions.

We had previously released an experimental static version of these embeddings for the contiguous United States, described in \citep{agarwal2024general}. We also release code recipes to easily integrate the embeddings for a wide range of use cases.\footnote{\url{https://github.com/google-research/population-dynamics}}
Furthermore, non-US static embeddings have been made available to research partners to evaluate performance and utility in public health tasks, like epidemiological modeling.

Second, we have now created a temporal version of the Population Dynamics Foundations embeddings. The initial version contains an embedding per month starting in July, 2023 up to the present. Each embedding is created in the same manner as the static Population Dynamics Foundations embedding, only with signals restricted to the given month. We computed a common vocabulary across all time points and geographic regions for the Search Trends data. Note that the temporal embeddings used in this report did not use the Knowledge Graph mapping used for the global version of the embeddings.

\paragraph{Experimental Labels}
Our original Population Dynamics Foundations benchmark \citep{agarwal2024general} used the following target variables from Google Data Commons and Google Earth Engine at the postal code and county level from 2022.
\begin{itemize}
    \item \textbf{21 Health:} High Cholesterol, Physical Health Not Good, Stroke, Binge Drinking, Physical Inactivity, Received Annual Checkup, Cancer (Excl. Skin Cancer), Diabetes, Mental Health Not Good, Coronary Heart Disease, High Blood Pressure, Received Cholesterol Screening, Received Dental Visit, Asthma, Chronic Kidney Disease, Arthritis, Chronic Obstructive Pulmonary Disease, High Blood Pressure (Medicated), Obesity, Sleep Less Than 7 Hours, and Smoking
    \item \textbf{6 Socioeconomic:} Median Income Household, Median Home Value, Night Lights, Population Density, Unemployment, and Poverty
    \item \textbf{2 Environmental:} Tree Cover and Elevation
\end{itemize}

The global embeddings in this work used four target variables from Google Earth Engine that overlap with our original set including Night Lights, Population Density, Unemployment, and Poverty. For the experiment training on a subset of European countries and evaluating on a held out country we used Nomenclature of Territorial Units for Statistics level 3 (NUTS 3) data \citep{Eurostat2024}. These labels included GDP per capita, Death rate, and Fertility rate. We used a Eurostat correspondence table\footnote{\url{https://ec.europa.eu/eurostat/web/nuts/correspondence-tables}} to map our postal code level embeddings to NUTS 3 level for modeling these labels.

For our temporal embedding analysis over time we used RSV, flu, and COVID emergency department visits datasets from the Population Health Information and Visualization Exchange (PopHIVE) \citep{pophive2025}. This dataset contained weekly emergency department visits data at the county level that we aggregated to monthly.

\subsection{Combining Earth AI Models: Predictive Applications}
\label{subsec:appendix_combining_earth_ai_models_predictive_applications}
\subsubsection{FEMA Risk Scores and CDC Health Statistics with Population Dynamics Foundations + AlphaEarth Foundations}
\label{subsubsec:appendix_fema_risk_scores_and_cdc_health_stats}
Leveraging the pretrained embeddings from AlphaEarth and Population Dynamics Foundations, we establish a simple framework to use embeddings as input features to predict any variable at the census tract level. This includes:
\begin{enumerate}
    \item \textbf{Aggregating AlphaEarth embeddings at the census tract level.} The AlphaEarth embeddings are available in Google Earth Engine as a yearly image collection with 64 bands per image, at 10m resolution. We selected the image from 2023 and converted each band into 10 features per census tract, for a total of 640 features per tract. The 10 features are derived as a fixed-sized histogram of the image band values.
    \item \textbf{Deriving Population Dynamics Foundations embeddings at the census tract level.} The Population Dynamics Foundations embeddings are natively at the US postal code level. We converted them to census tract level features via area-weighted averaging, based on 2020 land area overlap data from census.gov.
    \item \textbf{Using gradient boosting decision trees to predict the target variables of interest,} such as the FEMA NRI risk scores, given the input embeddings.
\end{enumerate}

The results above are modeled with gradient-boosted decision trees with a fixed set of hyperparameters, suggesting that a simple modeling setup can benefit from the gains of combining embeddings without tuning to each individual label.

\subsubsection{Disaster Relief with Google's Experimental Cyclones Forecast}
\label{subsubsec:appendix_disaster_relief_with_googles_experimental_cyclones_forecast}
Bellwether developed a model which predicts damage caused by hurricane winds. Specifically, the model classifies if buildings will be damaged or not given the underlying weather and build environment conditions.

\paragraph{Data:} We use a dataset containing over 1.5 million data points consisting of tens of thousands of residential and commercial properties across the United States. The dataset spans 6 years (2017-2022) and 22 hurricane events. It contains labels of damage/no damage and property-level information (e.g., building material, roof type). Six of the most recent events are held out for validation and testing. Data splits are designed such that distributions of important features (e.g. wind speeds) are relatively similar across training, validation, and testing. We supplement this data with hurricane information (e.g. track path, minimum pressure, max wind speed) from HURDAT2 \citep{landsea2013atlantic}, local environmental insights (e.g. building height from Google’s Open Buildings dataset, and Google’s Population Dynamics Foundations embeddings), and Google’s Experimental Cyclone model for hurricane track and wind speed forecast.

\paragraph{Training:} We trained a gradient boosted decision tree model on this data. Performance metrics on the held out test set are displayed below. Predicting hurricane damage is made difficult by heavy class imbalance. Achieving high accuracy (96\% in our case) overall is not difficult due to the fact that the overwhelming majority of buildings in a given dataset will usually remain undamaged. Notably, the important metrics are the recall, precision, and F1 score of the minority damaged building class (respectively 0.6, 0.57, and 0.59 in our model).

\begin{table}[H]
\centering
\caption{Performance metrics (Recall, Precision and F1 score) per class on a held out test set. These are computed for model predictions using observed hurricane tracks from HURDAT2.}
\label{tab:app_b1}
\resizebox{0.6\textwidth}{!}{%
\begin{tabular}{lccc}
\toprule
& Recall & Precision & F1 score \\ \midrule
Class 0: Undamaged Buildings & 0.98 & 0.98 & 0.98 \\
Class 1: Damaged Buildings & 0.60 & 0.57 & 0.59 \\ \bottomrule
\end{tabular}%
}
\end{table}

\paragraph{Forecasting:} We leveraged this trained model to predict hurricane damage days in advance of the actual hurricane as a tool to forecast disaster relief needs. For this, we utilize forecasted hurricane tracks from Google’s experimental cyclone model instead of historically observed ones as the input to our model. Static features such as Population Dynamics Foundations insights and building information remain the same.

In Figure~\ref{fig:cyclone_damage} we show the performance of the model 3 days in advance of Hurricane Ian landfall (2022). The Google Experimental Cyclone model provides 50 ensemble members, and each ensemble member is run through our trained model, generating a distribution of damaged buildings which are compared to the actual number of damaged buildings. Distributions of precision, recall, and F-1 score are also assessed for each of the 50 ensemble members.

\paragraph{Notes and limitations:}
\begin{itemize}
    \item Figure~\ref{fig:cyclone_damage} only showed results for one hurricane and one lead time at the moment. We plan to expand these results to have different lead times (i.e., not just 3 days) and different hurricanes.
    \item In addition, the analyzed dataset represents a subset of all buildings within the affected hurricane zone. The true number of damaged properties by Hurricane Ian would be significantly greater.
    \item Per hurricane event, less than 4\% of properties are labeled as damaged across the entire dataset, displaying heavy data imbalance.
    \item The presence of mislabeled data (e.g. a small portion of buildings may be damaged due to floods and not because of wind damage) might also introduce noise and hinder model performance.
\end{itemize}

\subsection{Geospatial Reasoning across Earth AI Models}
\label{subsec:appendix_geospatial_reasoning_across_earth_ai_models}
\subsubsection{Agentic Framework}
\label{subsubsec:appendix_agentic_framework}
This research employed an agentic framework using the Agent Development Kit (ADK) \citep{google2025adk}, featuring Gemini 2.5 Flash and Pro models.

\paragraph{Domain-oriented Architecture:} Beyond general orchestration, planning, and recovery, we organized our geospatial functionality by domain; some domains succeeded as a simple list of tools, while others needed more steering and guidance for the tasks at hand, which we encapsulated in “expert” agents.

\begin{description}
    \item[Demographics (Expert):] Selects and retrieves optimal statistical variables from Data Commons \citep{DataCommons2025} to answer natural language statistical queries.
    \item[Weather (Expert):] Fetches and interprets both historical \citep{gorelick2017google, hersbach2020era5} and forecast weather data, including precipitation, weather-related information such as temperature \citep{google2025weather}, wind speed and cyclones \citep{google2025weatherlab}, and floods \citep{nearing2024global, nevo2022flood}.
    \item[Spatiotemporal Model Training (Expert):] Offers a high-level interface for advanced modeling tasks like spatial interpolation, super-resolution, and forecasting by training lightweight models on the fly with pre-trained embeddings, powered by foundation models like Population Dynamics Foundations \citep{agarwal2024general} and TimesFM \citep{das2024decoderonly}.
    \item[Earth Engine (Expert):] Identifies and fetches the optimal Google Earth Engine \citep{gorelick2017google} datasets and data bands for a user's query by programmatically searching the public data catalog using Google Search.
    \item[Search (Expert):] A generalist agent that uses Vertex AI Grounding with Search as a fallback knowledge source when more specialized tools lack the necessary information.
    \item[Location and Places (Toolset):] Resolves various types of geospatial entities—from broad administrative areas like countries and counties to specific points of interest such as train stations or hospitals—into precise polygonal geometries using tools like Google’s Places API \citep{google2025places} and Places Aggregate API \citep{google2025placesaggregate}. This conversion is foundational for conducting spatial analysis, allowing the system to reason about geographic boundaries and intersections, such as calculating the overlap between a county and a cyclone's trajectory.
    \item[Remote Sensing and Imagery (Toolset):] Employs RemoteSensing Foundation models \citep{barzilai2025recipe} for open vocabulary image classification, retrieval and object detection. It utilizes a split-processing approach to cache image embeddings, which significantly accelerates repeated analysis of the same areas.
\end{description}
This domain-oriented architecture will further allow us to more easily “plug-in” new geospatial domains as we continue to develop the agent, as well as let us easily switch how a domain is implemented as LLMs and frameworks advance

\paragraph{Reasoning:} Gemini’s inherent ability to reason is fairly developed when provided all relevant data up-front and that data is natural language in nature. Where it needs help is with access to specialized data, specialized capabilities, special handling of non-language data (highly structured and/or numeric-heavy, such as geometry), and then more advanced orchestration and planning to weave everything together properly. This last aspect becomes more critical the more tools and capabilities the agent has, as well as the more complex it is to find and join the relevant data.

\paragraph{User Interface:} The agent is designed to be employed with a map-based user interface capable of visualizing geospatial results (e.g., pins, Earth Engine tile layers, and colored regions) in tandem with textual responses. While this UI was not part of the formal benchmark, it was crucial for the development process and is essential for practical applications.

\subsubsection{Baseline Gemini Configuration}
\label{subsubsec:appendix_baseline_gemini_configuration}
For our comparative analysis, we established baselines using the Gemini 2.5 Pro and Flash models implemented as Agent Development Kit (ADK) agents with the following Gemini built-in tools enabled to allow for online data access and code-based analysis: Grounding with Google Search, Grounding with Google Maps, and Code Execution. No custom tools were added. Additionally, to encourage these agents to always provide an answer, rather than abstaining from answering, as well as supply evidence of the sources used during reasoning, we included the following instructions in our Gemini agent prompt:
\begin{quote}
    Do your best to answer the user's question with the tools provided. If definitive answers can't be found, provide a best estimate. Cite sources used for the response and any reasoning used.
\end{quote}
Finally, we updated our rubric-based evaluation to give ``partial credit'' when the model attempted a response using available data, even if it did not perfectly match the prompt's intent.

\subsubsection{Benchmark Generation and Eval Metrics}
\label{subsubsec:appendix_benchmark_generation_and_eval_metrics}
\phantomsection
\paragraph{Fact-finding and Data Analytics}
\label{para:fact_finding}
Prompts were selected to encompass a comprehensive spectrum of capabilities across established benchmark categories, including demographics, geographical locations, historical meteorological data, and instances necessitating the synergistic interaction of multiple tools. A subject matter expert in geospatial analysis then rigorously reviewed all benchmark questions to confirm they were generalizable, useful for analysts, and relevant to the operational functions of our agents and tools.

\begin{table}[H]
\centering
\caption{A question and answer pair from each data domain and semantic category.} \label{tab:app_c1}
\resizebox{\textwidth}{!}{%
\begin{tabular}{p{0.2\textwidth} p{0.4\textwidth} p{0.4\textwidth}}
\toprule
& \textbf{Descriptive and Retrieval} & \textbf{Analytical and Relational} \\ \midrule
\textbf{Places} & Q: How many pharmacies are there in Portland, ME? Respond by filling in the blanks: Pharmacies: \_\_ \newline \newline A: Pharmacies: 21 & Q: What percentage of all restaurants in NYC are Korean restaurants? Respond by filling in the blanks: Korean restaurants: \_\_, Total restaurants: \_\_, Percentage (0.00): \_\_\% \newline \newline A: Korean restaurants: 364, Total restaurants: 23243, Percentage (0.00): 1.57\% \\
\midrule
\textbf{People and Communities} & Q: Respond by filling in the blanks: In Illinois in 2022, the diabetes prevalence was \_\_\% and the asthma prevalence was \_\_\% \newline \newline A: In Illinois in 2022, the diabetes prevalence was 10.63\% and the asthma prevalence was 9.60\% & Q: In 2020, which county in the US with an above average population had the highest percentage of households without home internet? Respond by filling in the blanks: County: \_\_, Households without internet (0.00): \_\_\% \newline \newline A: County: San Juan Municipio, Households without internet (0.00): 36.70\% \\
\midrule
\textbf{Weather and Environment} & Q: Which day in July 2024 was the hottest in New York City? Respond by filling in the blanks: Date (YYYY-MM-DD): \_\_, Temperature: \_\_ °F \newline \newline A: Date (YYYY-MM-DD): 2024-07-16, Temperature: 95.7 °F & Q: Compute the average elevation and max temperature (era5 land in Earth Engine) in July 2024 for counties in Arizona. What's the correlation between the temperature and elevation? Respond by filling in the blanks: Correlation (0.00): \_\_, County with the highest temperature: \_\_ \newline \newline A: Correlation (0.00): -0.87, County with the highest temperature: La Paz County \\
\midrule
\textbf{Multiple Domains} & Q: Which Florida county with population over 200K had recorded the max amount of rain on September 27th 2024 according to era5 land daily in Earth Engine? Respond by filling in the blanks: County: \_\_, Rain (0.00): \_\_ inches \newline \newline A: County: Leon County, Rain (0.00): 2.32 inches & Q: Identify the 5 Michigan counties with the most primary school students in 2023 and calculate the average number of students per school in each of these counties. Which county had the highest number of students per school and what was the number? Round to the nearest integer and respond by filling in the blanks: County: \_\_, Students per school: \_\_ \newline \newline A: County: Wayne County, Students per school: 446 \\
\bottomrule
\end{tabular}%
}
\end{table}

To create ground truth answers for scoring, we first generated answers for each prompt using direct API calls or dataset queries. These answers were then verified and, if necessary, corrected by geospatial experts. All questions in the benchmark are fill-in-the-blanks. The scores are computed by comparing the values that the agent filled in against the values in the gold answer, while ignoring all surrounding text. Textual values are scored using ROUGE-L F1 \citep{lin-2004-rouge} and numerical values are scored based on a clamped absolute percentage error. For each answer, all scores are averaged to obtain the final instance score. In order to account for data source differences, businesses opening and closing over time, and uncertainties in statistical data, numerical values in the answer that are within +/-10\% of the golden values are given a score of 1.0. Values that differ by more than 100\% from the golden answer are given a score of 0.0. The rest of the values are assigned a score of 1.0 minus the absolute percentage error.

\phantomsection
\paragraph{Crisis Response Case Study}
\label{para:crisis_response_case_study}
Our crisis response prompts were specifically designed to test the full suite of the agent's predictive capabilities, including RemoteSensing Foundation, Population Dynamics Foundations, and meteorological, cyclone, flood, and TimesFM forecasting. Each prompt was formulated to address a complex crisis response scenario that required the agent to synthesize data from a predictive model with geospatial or demographic information to generate actionable insights.

This benchmark was a deliberate shift from prior research, which often tested these models in isolation. Our primary goal was to evaluate the agent's ability to reason effectively within the context of a crisis, using its entire predictive modeling capabilities.

To that end, we designed a custom set of rubrics for each scenario which scrutinized reasoning and data characteristics of the agent's response. For instance, with a non-deterministic task like weather forecasting, our rubrics verified the logical process behind the result (e.g., ensuring the agent forecasting in the requested areas of interest and timeframe). We also qualitatively verified the final output for coherence, confirming that it delivered the final answer in the requested format (e.g., a list of areas versus a total count of areas).

\begin{table}[H]
\centering
\caption{The crisis response prompts; IDs are from Table~\ref{tab:crisis_response} that provides an overview of the crisis response benchmark results.} \label{tab:app_c2}
\resizebox{0.78\textwidth}{!}{%
\begin{tabular}{lp{0.85\textwidth}}
\toprule
\textbf{ID} & \textbf{Prompt} \\ \midrule
1 & Help me assess the flood risk to agricultural infrastructure in Scott County, Minnesota, due to predicted flooding as of August 20th, 2025 at 12:00 a.m. For the high risk flood prediction in that county, provide a table that lists each impacted zip code and a count of farms for each one. Don't include zip codes without farms, and sort the table by zip code ascending. \\
\addlinespace
2 & I would like to evaluate the risk of flooding on vulnerable populations in Matanuska-Susitna Borough, Alaska, due to forecasted flooding on September 2nd, 2025 (as of 12:00 A.M. local time in Anchorage, AK). Using flood risk data, identify at risk zip codes that have a large percentage of vulnerable populations. Please help me by: (1) Identifying all of the zip codes that should be notified of floodings, and (2) for each identified zip code, producing the average income of the people living there, as well as the percentage of people over 60 years of age (out the the entire population of that zip code). Format this as a table. \\
\addlinespace
3 & For upcoming Typhoon Ragasa as of September 18, 2025 12:00 UTC, list the administrative regions in the Philippines that were forecast to experience gale-force winds in the next seven days. \\
\addlinespace
4 & Which are the top 5 most populated counties of Florida that were in the forecasted cyclone path of Hurricane Helene as of September 25, 2024 12:00 UTC? Respond with a table of each county with its population, sorted by county name ascending. \\
\addlinespace
5 & Help me assess the impact of a widespread drought across Texas by identifying the five counties with the most vulnerable farms based on normalized vegetation index, estimating each county’s 10-day rainfall forecast and determining which ones are unlikely to receive meaningful precipitation or drought relief during the forecast period. When listing the counties, include their NVI and their 10-day rainfall forecast. \\
\addlinespace
6 & I am a State emergency manager in New Mexico and I am planning to prioritize the establishment of cooling centers for the upcoming heat wave. For counties that have a total population greater than 50,000, find the top 5 most vulnerable ones ( percentage of population over 70 ) that are forecasted to experience a daily high temperature over 80 degrees F on any of the next 5 days. \\
\addlinespace
7 & Following the recent wildfire in the Island of Maui, assess the scope of residential damage in the town Lahaina. Find and directly interpret satellite imagery tiles that were taken after the disaster (prefer earliest available images) to classify areas that were damaged by the fire and identify which schools and grocery stores in the city were located in those areas. List how many tiles were analyzed and which percentage of them showed damage. \\
\addlinespace
8 & I'm concerned about potential flooding from lakes in residential areas in Wakulla County, Florida. For the most populous zip code in the county, use the latest satellite imagery, divide it between residential, agricultural and natural landscape, and report on the percentage of each. In the residential areas, how many lakes are detected? Provide a table of the zip codes in the county and their population. \\
\addlinespace
9 & Hurricane Helene at the end of September 2024 caused widespread damage in its path. I would like to assess the effect of this storm on unemployment rates. Use the data from January to September 2024 to forecast the unemployment rate for October 2024 in all US counties, and compute the difference between the actual values and forecasted values for that month. List the top 5 counties by positive difference, and for each county, provide the actual unemployment rate, the forecasted unemployment rate, and the difference between the two. Were all of those counties affected by Helene according to a web search? \\
\addlinespace
10 & Buncombe County was one of the hardest hit places by hurricane Helene. What are the postal code level unemployment rates in the month immediately following the hurricane in this county? \\
\bottomrule
\end{tabular}%
}
\end{table}

\begin{longtable}{lp{0.85\linewidth}}
\caption{The crisis response rubrics; IDs are from Table~\ref{tab:crisis_response}. crisis response benchmark results} \label{tab:app_c3} \\
\toprule
\textbf{ID} & \textbf{Rubrics} \\ 
\midrule
\endfirsthead

\multicolumn{2}{c}{{\tablename\ \thetable{} -- Continued from previous page}} \\
\toprule
\textbf{ID} & \textbf{Rubrics} \\ 
\midrule
\endhead

\midrule
\multicolumn{2}{r}{{Continued on next page}} \\
\endfoot

\bottomrule
\endlastfoot

1 & \begin{itemize} \item The response must provide information that authoritative flood data was used. Partial credit is allowed for non-authoritative data. \item The response must provide information that authoritative data was used to find farms. Partial credit is allowed for non-authoritative data. \item The response must provide information that there was a process to precisely identify which zip codes to include in the response that used flood data. \item A response based on estimates or approximations around missing data is allowed partial credit but not full credit. \item The response must provide information that the zipcodes are from Scott County, Minnesota \item The response must provide an answer to the question and not just be a plan of how someone could answer the question \item The response must have a list of zip codes impacted. \item The response must contain a count of farms per zip code \end{itemize} \\
\addlinespace
2 & \begin{itemize} \item The response must provide information that authoritative flood data from a data source was used and that flood data was available for the area. \item The response must provide information that authoritative statistics data from a data source was used to get information on the percentage of people over 60 (or alternatively on total population and total population over 60). \item The response must provide information that there was a process to precisely identify which zip codes to include in the response that used flood data. \item The response must not be based on estimates or approximations around missing data. \item The response must produce more than one zip code, but not include all zip codes in the borough. \item The response must have a list of zip codes, each with a specific average income and a specific percentage of people over 60. \end{itemize} \\
\addlinespace
3 & \begin{itemize} \item The response must provide information that authoritative cyclone forecasting data from a data source was used. Partial credit is allowed for non-authoritative data. \item The response must provide information that authoritative statistics data from a data source was used to get information on regions in the Philippines. Partial credit is allowed for non-authoritative data. \item The response must provide information that there was a process to precisely identify which regions to include in the response that used cyclone data. \item The response must provide information that gale force winds information was used to identify impacted regions in the Philippines. \item A response based on estimates or approximations around missing data is allowed partial credit but not full credit. \item The response must produce more than one region, but not include all the regions in the state; violating this should still give partial credit. \item The response must provide an answer to the question and not just be a plan of how someone could answer the question. \item The response must have a list of regions impacted. \end{itemize} \\
\addlinespace
4 & \begin{itemize} \item The response must provide information that authoritative cyclone forecasting data from a data source was used. Partial credit is allowed for non-authoritative data. \item The response must provide information that authoritative statistics data from a data source was used to get information on the population of counties in Florida (or alternatively counties in Florida, and population therein). Partial credit is allowed for non-authoritative data. \item The response must provide information that there was a process to precisely identify which administrative areas to include in the response that used cyclone data. \item A response based on estimates or approximations around missing data is allowed partial credit but not full credit. \item The response must have at least one county, but if it has less than 5 it gets partial credit. \item The response must provide an answer to the question and not just be a plan of how someone could answer the question. \item The response must have a list of counties impacted. \item The response must contain the population for each county listed. \end{itemize}
\\
\addlinespace
5 & \begin{itemize} \item The response must provide information that authoritative weather data was used. Partial Credit is allowed for non-authoritative data. \item The response must provide information that authoritative data was used for counties and NVI. Partial Credit is allowed for non-authoritative data. \item Must provide a list of at most 5 counties in Texas with vulnerable farms based on normalized vegetation index (NVI) along with their 10-day precipitation forecast and their assessment of drought relief. \item The response must provide information about the process used to identify vulnerable counties as well as drought assessment. \item A response based on estimates or approximations around missing data is allowed partial credit but not full credit. \end{itemize} \\
\addlinespace
6 & \begin{itemize} \item The response must provide information that an authoritative data was used for temperature forecasts and population demographics. \item The response must provide a list of at most 5 counties in New Mexico that meet the criteria. \item The response must provide information that there was a process using demographic data and weather forecasts to precisely identify which counties to include in the response. \end{itemize} \\
\addlinespace
7 & \begin{itemize} \item The response must provide information that Lahaina, Maui satellite imagery was used and that it was taken at the earliest date available after the disaster. For full credit, the response must have evidence that image tiles were directly analyzed and that the information does not come from another source's report of satellite image tile analysis. If satellite imagery tiles are not directly analyzed but another authoritative data source is found for discovering residential damage, partial credit is allowed. \item The response must provide information that authoritative places data from a data source was used to get information on schools and grocery stores locations. \item The response must provide information that damaged areas were identified in the town of Lahaina using satellite imagery. \item The response must provide information that there was a process to precisely identify which schools and grocery stores were impacted based on residential damage. \item A response based on estimates or approximations around missing data is allowed partial credit but not full credit. \item The response must produce more than one school and one grocery store, but not include all schools and grocery stores in Lahaina; violating this should still give partial credit. \item The response must provide an answer to the question and not just be a plan of how someone could answer the question. \item The response must have a list of impacted schools and grocery stores in Lahaina. \item The response must have a count of how many tiles were analyzed and the percentage of those tiles that showed damage; if tiles were not analyzed only give partial credit. \end{itemize} \\
\addlinespace
8 & \begin{itemize} \item The response must provide information that satellite imagery was used and that it was recent imagery. Derived information from satellite imagery can get partial credit, but not full credit. \item The response must provide information that authoritative statistics data from a data source was used to get population information. \item The response must provide information that a process was used to precisely identify the most populous zip code based on population data. \item The response must provide information that the selected zip code was analyzed for area classification; only three classifications are allowed, any other classes reduces the score but still grants partial credit. \item The response must provide information that lakes were identified in the residential areas and not other areas. \item The response must provide an answer to the question and not just be a plan of how someone could answer the question. \item The response must have the number of lakes detected in the residential areas in that zip code as well as the percentage of each area classification in that zip code. \end{itemize} \\
\addlinespace
9 & \begin{itemize} \item The agent must identify the Bureau of Labor Statistics (BLS) as the source for unemployment data. \item The agent must list exactly five counties, all of which must be located in North Carolina, and provide the actual and forecasted unemployment values for each. \item The agent must provide the difference between the actual and forecasted unemployment values for each county. \item The agent must confirm that all five listed counties experienced significant damage from the storm. \end{itemize} \\
\addlinespace
10 & \begin{itemize} \item Should identify that Hurricane Helene occurred in September 2024. \item Should identify that the month of interest for unemployment rates is October 2024. \item Should attempt to retrieve unemployment data at the zip code level before resorting to a model; any statement around data being unavailable is acceptable. \item Should retrieve county-level unemployment data. \item Should identify and cite the source of the county-level data. \item Should use a model to estimate the zip code level unemployment rates from the retrieved county-level data. \item Should provide a final answer that includes a list of 15 zip codes and their corresponding estimated unemployment rates. \end{itemize} \\
\end{longtable}

\phantomsection
\paragraph{Limitations with rubrics}
\label{para:limitations_with_rubrics}
As stated earlier, the use of rubrics allowed the evaluation to focus on the agent's reasoning instead of the performance of each of the predictive models. For example, unlike in the Q\&A benchmark, we could evaluate whether the agent had a logical process and was using appropriate data sources. However, this method revealed its own challenges:
\begin{itemize}
    \item \textbf{Inconsistent Outputs from Consistent Reasoning when Using Predictive Models:} An agent can follow a consistent reasoning process, yet produce different final answers when the outputs from its underlying predictive models vary. In the Maui fire scenario, for example, the agent's logic was stable, but the final list of impacted schools and stores changed whenever the RemoteSensing Foundation model identified a different set of fire-damaged tiles.
    \item \textbf{Difficulty in Verifying Methodology:} It is challenging with rubrics to confirm whether the agent was performing a technical geospatial task itself (like analyzing satellite imagery) or simply synthesizing pre-existing information from sources like Google Search or its parametric knowledge. This was particularly evident with high-profile events with extensive public coverage, where an agent's response might imply direct data analysis, but further prompt refinement could reveal it was actually synthesizing secondary sources of information such as search results and news articles.
\end{itemize}
The first challenge can be addressed by the use of “gold answers” to measure output correctness. To address the second challenge, future work will aim to further improve rubrics but also combine them with human expert review to leverage multiple perspectives and reduce reliance on a single source of score

\paragraph{Example responses}
The following is an example of our Agent and Gemini answering the second question (ID 2). For each, a representative response is chosen from the responses with a median score.

\paragraph{Prompt:}
\begin{quote}
I would like to evaluate the risk of flooding on vulnerable populations in Matanuska-Susitna Borough, Alaska, due to forecasted flooding on September 2nd, 2025 (as of 12:00 A.M. local time in Anchorage, AK). Using flood risk data, identify at risk zip codes that have a large percentage of vulnerable populations. Please help me by: (1) Identifying all of the zip codes that should be notified of floodings, and (2) for each identified zip code, producing the average income of the people living there, as well as the percentage of people over 60 years of age (out of the the entire population of that zip code). Format this as a table.
\end{quote}

\paragraph{Agent Response:}
Score breakdowns: 8 x 1.0; 2 x 0.0

Here are the zip codes in Matanuska-Susitna Borough, Alaska, that are at risk of flooding due to the forecasted flood on September 2nd, 2025, along with their average household income and the percentage of the population over 60 years of age:

\begin{center}
\begin{tabular}{c c c}
\toprule
\textbf{Zip Code} & \textbf{Mean Household Income} & \textbf{Percentage Over 60} \\
\midrule
99667 & \$52239 & 69.08\% \\
99683 & \$88328 & 53.08\% \\
99688 & \$83749 & 33.20\% \\
\bottomrule
\end{tabular}
\end{center}


\textbf{Reasoning:}
\begin{enumerate}
    \item Flood forecast data for Matanuska-Susitna Borough, Alaska, for September 2nd, 2025, was obtained.
    \item A list of all US postal codes (zip codes) within Matanuska-Susitna Borough, Alaska, was retrieved.
    \item The polygon geometries for these zip codes were fetched.
    \item A spatial intersection was performed between the forecasted flood impact area and the zip code polygons to identify the zip codes that are at risk of flooding.
    \item For each identified at-risk zip code, the average household income and the percentage of the population over 60 years of age were retrieved.
\end{enumerate}

\textbf{Data Provenance:}
\begin{itemize}
    \item Flood Forecasts: Google Flood Forecasts
    \item Mean Household Income: \url{https://data.census.gov/cedsci/table?q=S1901&tid=ACSST5Y2019.S1901} (2023)
    \item Count of People 60 Or More Years: \url{https://data.census.gov/table?q=S0101:+Age+and+Sex&tid=ACSST1Y2022.S0101} (2023)
    \item Count of Person: \url{https://www.census.gov/programs-surveys/acs/data/data-via-ftp.html} (2023)
\end{itemize}

\paragraph{Gemini Response:}
Score breakdowns: 1 x 0.75; 4 x 0.5; 5 x 0.0

Based on the flood risk data for September 2nd, 2025, in the Matanuska-Susitna Borough, Alaska, the following zip codes are identified as being at risk of flooding. The table below provides the average income and the percentage of the population over 60 years of age for each of these zip codes to help assess the vulnerability of the population.

The flood forecast indicates that the Western, Central, and Northern Susitna Valley were under a flood advisory. The zip codes within these areas are therefore considered at risk.

\textbf{At-Risk Zip Codes and Vulnerable Population Analysis}

\begin{center}
\begin{tabular}{ccccc}
\toprule
\textbf{Zip Code} & \textbf{Location(s)} & \textbf{Average Income} & \textbf{Percentage of...} & \textbf{Notes} \\ \midrule
99667 & Skwentna & \$82,744 & >50\% & Located in... \\
99683 & Willow/Trapper Creek & \$88,328 & \textasciitilde52.6\% & Located in... \\
99674 & Sutton & \$90,385 & \textasciitilde26.4\% & Located in... \\
99645 & Palmer & \$118,167 & 19\% & Located in... \\
99652 & Big Lake & \$96,500 & \textasciitilde13.9\% (over 65) & Located in... \\
99676 & Talkeetna & \$86,473 & >9.4\% & Located in... \\
99654 & Wasilla & \$115,548 & 6.7\% & Located in... \\
99694 & Houston & \$72,411 & \textasciitilde6\% & Located in... \\ \bottomrule
\end{tabular}
\end{center}


\textbf{Summary of Findings and Recommendations}

Based on the analysis, the following zip codes should be prioritized for flood notifications and support due to a combination of flood risk and a vulnerable population (a high percentage of residents over 60 and, in some cases, lower income levels):
\begin{itemize}
    \item \textbf{99667 (Skwentna):} With over 50\% of the population over 60, this area is highly vulnerable.
    \item \textbf{99683 (Willow/Trapper Creek):} Similar to Skwentna, this area has a very high percentage of residents over 60 (around 52.6\%).
    \item \textbf{99674 (Sutton):} A significant portion of the population is over 60 (around 26.4\%), and the senior population is described as ``extremely large''.
    \item \textbf{99676 (Talkeetna):} While the precise percentage of the population over 60 is not available, it is described as ``extremely large'', indicating a potentially vulnerable population.
\end{itemize}
These areas have a combination of high flood risk and a significant population of older adults who may require additional assistance during a flooding event.

\textbf{Limitations:}
\begin{itemize}
    \item The demographic data is based on publicly available information and may have some variations depending on the source.
    \item For some zip codes, the percentage of the population over 60 was not explicitly available and was estimated based on other demographic indicators.
    \item There were some discrepancies in zip code numbers for certain locations between different data sources. The analysis used the most detailed and geographically relevant data available.
    \item Demographic data was not available for all at-risk zip codes identified in the flood-prone areas, specifically 99623 (Big Lake area), 99629 (Houston area), 99687 (Wasilla area), and 99688 (Susitna area). These areas are also at risk of flooding and should be included in any notification efforts.
\end{itemize}

\paragraph{Autorater Rubrics:}
\begin{itemize}
    \item The response must provide information that authoritative flood data from a data source was used and that flood data was available for the area.
    \item The response must provide information that authoritative statistics data from a data source was used to get information on percentage of people over 60 (or alternatively on total population and total population over 60).
    \item The response must provide information that there was a process to precisely identify which zip codes to include in the response that used flood data.
    \item The response must not be based on estimates or approximations around missing data.
    \item The response must produce more than one zip code, but not include all zip codes in the borough; violating this should still give partial credit.
    \item The response must have a list of zip codes, each with a specific average income and specific percentage of people over 60.
\end{itemize}

\paragraph{Autorater Response (Agent):}
\texttt{“score”: COMPLETELY (1.0)}

\texttt{“answer”:}
\begin{enumerate}
    \item The response explicitly states that 'Flood forecast data for Matanuska-Susitna Borough, Alaska, for September 2nd, 2025, was obtained' and lists 'Google Flood Forecasts' as a data provenance, fulfilling the requirement for using authoritative flood data and its availability.
    \item The response lists 'Mean Household Income: \url{https://data.census.gov/cedsci/table?q=S1901&tid=ACSST5Y2019.S1901} (2023)', 'Count of People 60 Or More Years: \url{https://data.census.gov/table?q=S0101:+Age+and+Sex&tid=ACSST1Y2022.S0101} (2023)', and 'Count of Person: \url{https://www.census.gov/programs-surveys/acs/data/data-via-ftp.html} (2023)' as data provenance, which are authoritative statistics data sources, fulfilling this requirement.
    \item The reasoning clearly outlines a process: 'A list of all US postal codes (zip codes) within Matanuska-Susitna Borough, Alaska, was retrieved. The polygon geometries for these zip codes were fetched. A spatial intersection was performed between the forecasted flood impact area and the zip code polygons to identify the zip codes that are at risk of flooding.' This precisely identifies the zip codes using flood data.
    \item The response provides specific values for income and percentage over 60 for each zip code, and the reasoning describes a process of retrieving this data, implying no estimates or approximations due to missing data were used.
    \item The response produces three zip codes (99667, 99683, 99688), which is more than one and not all zip codes in the borough, fulfilling this requirement.
    \item The response provides a table with a list of zip codes, each with a specific average income and specific percentage of people over 60, fulfilling this requirement.
\end{enumerate}

\paragraph{Autorater Response (Gemini):}
\texttt{“score”: MODERATELY (0.5)}

\texttt{“answer”:}
\begin{enumerate}
    \item The response does not explicitly state that authoritative flood data from a specific data source (like Google flood data) was used. It mentions 'flood risk data' and 'flood forecast' but doesn't name a source.
    \item The response does not explicitly state that authoritative statistics data from a specific data source was used for demographic information. It mentions 'publicly available information' but doesn't name a source.
    \item The response describes a process for identifying zip codes based on flood advisories in specific valleys (Western, Central, and Northern Susitna Valley), which implies the use of flood data to determine at-risk areas.
    \item The response explicitly states that for some zip codes, 'the percentage of the population over 60 was not explicitly available and was estimated based on other demographic indicators.' It also notes that 'Demographic data was not available for all at-risk zip codes identified in the flood-prone areas.' This violates the criterion that the response must not be based on estimates or approximations around missing data.
    \item The response provides 8 zip codes, which is more than one but not all zip codes in the borough, fulfilling this criterion.
    \item The response provides a table with a list of 8 zip codes, each with a specific average income and a specific percentage of people over 60 (though some are estimates or ranges, which ties into point 4).
\end{enumerate}

The primary reason for the 'Moderately' rating is the explicit admission of using estimates and having missing data for some zip codes, which directly contradicts a key requirement. Additionally, the lack of explicit naming of authoritative data sources for both flood and demographic data reduces the quality of the answer.

For the 10-run mean of this prompt, our Geospatial Reasoning Agent significantly outperformed baseline Gemini 2.5 Pro, achieving an average Likert score of 0.80 compared to Gemini's 0.28 over ten trials. This outcome underscores the value of providing direct access to geospatial data and capabilities: the baseline Gemini agent must rely on secondary sources via Google Search, such as reports and news articles that may lack completeness or depth, while our agent is equipped to interact with and analyze the primary data source and the flood forecasting model, leading to a more robust and reliable assessment.

\subsubsection{Automated result evaluation for Crisis Response}
\label{subsubsec:appendix_automated_result_evaluation_for_crisis_response}
For prompts in the Crisis Response Case Study section, answers are free-form; we use an autorater based on prompting Gemini 2.5 Flash for evaluating aspects of the answers (with the aspects/rubrics defined together with the questions in the section above.) The following autorater prompt template was used:
\begin{quote}
    You are an expert at evaluating the quality of an answer to a user request (question/prompt) about geospatial data asked by a user and answered by a geospatial agentic information system.

    For each criterion listed, output an explanation for why the candidate answer fulfills or does not fulfill that specific criterion; make a numbered list of these explanations. After considering all of the listed criteria, use the combination of all of these explanations to determine the final rating.
    If the candidate answer contains a computed value and the rating criteria specify a range for that value, the candidate answer must contain a value that is within the specified range.
    If a rating criterion specifies an answer strategy, the candidate answer must reference using that strategy (i.e. looking up a specific type of value and performing specified calculations).
    Authoritative data sources are data sources that are considered to be authoritative for the subject matter of the user request. In addition to data directly from governmental agencies such as the CDC and other well-known organizations such as the World Bank, the following are examples of authoritative data sources for respective domains:
    \begin{itemize}
        \item Flood: Google Floods data.
        \item Statistics and Demographics: Data Commons.
        \item Geospatial Information:Google Maps, Places API, Google Earth Engine.
        \item Cyclones: Google Cyclones Forecast and Tracking data, Weather Lab.
        \item Weather: Google Weather data, and Google Maps Platform Weather API.
    \end{itemize}

    User request:
    \textless{}question text\textgreater{}

    Agent response:
    \textless{}response text\textgreater{}

    Rating criteria for the answer fulfilling the requirements:
    \textless{}question-specific rating criteria\textgreater{}
\end{quote}

In addition the model prompt for autorating defined a five-point Likert-style rating scale:
\begin{quote}
    Your task is to determine whether the candidate answer fulfills the requirements specified in the rating criteria.
    \begin{description}
        \item[not\_at\_all] = The candidate answer completely fails to fulfill the requirements specified in the rating criteria.
        \item[somewhat] = The candidate answer contains some information aligned with the rating criteria, but is missing some specified key requirements.
        \item[moderately] = The candidate answer fulfills a significant portion of the requirements specified in the rating criteria but misses out on enough of the requirements that the answer cannot be considered satisfactory.
        \item[mostly] = The candidate answer generally fulfills the requirements specified, such that it can be considered mostly satisfactory, but misses some details or nuances.
        \item[completely] = The candidate answer fulfills all of the requirements specified in the rating criteria.
    \end{description}
\end{quote}
These ratings were mapped to numerical values [0, 0.25, 0.5, 0.75, 1] for the combined metrics computation.

\end{document}